# A review of motion planning algorithms for intelligent robotics

Chengmin Zhou[1]           {czhou@uef.fi}
Bingding Huang[2]          {huangbingding@sztu.edu.cn}
Pasi Fränti[1,*]           {franti@cs.uef.fi}

[1] University of Eastern Finland, Joensuu, Finland.
[2] Shenzhen Technology University, Shenzhen, China.
* Correspondence: Pasi Fränti

**Abstract:** We investigate and analyze principles of typical motion planning algorithms. These include traditional planning algorithms, supervised learning, optimal value reinforcement learning, policy gradient reinforcement learning. Traditional planning algorithms we investigated include graph search algorithms, sampling-based algorithms, and interpolating curve algorithms. Supervised learning algorithms include MSVM, LSTM, MCTS and CNN. Optimal value reinforcement learning algorithms include Q learning, DQN, double DQN, dueling DQN. Policy gradient algorithms include policy gradient method, actor-critic algorithm, A3C, A2C, DPG, DDPG, TRPO and PPO. New general criteria are also introduced to evaluate performance and application of motion planning algorithms by analytical comparisons. Convergence speed and stability of optimal value and policy gradient algorithms are specially analyzed. Future directions are presented analytically according to principles and analytical comparisons of motion planning algorithms. This paper provides researchers with a clear and comprehensive understanding about advantages, disadvantages, relationships, and future of motion planning algorithms in robotics, and paves ways for better motion planning algorithms.

## I. Introduction

Intelligent robotics, nowadays, is serving people from different backgrounds in complex and dynamic shopping malls, train stations and airports [1] like Daxin in Beijing and Changi in Singapore. Intelligent robots guide pedestrians to find coffee house, departure gates and exits via accurate motion planning, and assist pedestrians in luggage delivery. Another example of intelligent robotics is parcel delivery robots from e-commercial tech giants like JD in China and Amazon in US. Researchers in tech giants make it possible for robots to autonomously navigate themselves and avoid dynamic and complex obstacles via applying motion planning algorithms to accomplish parcel delivery tasks. In short, intelligent robotics gradually play a significant role in service industry, agricultural production, manufacture industry and dangerous scenarios like nuclear radiation environment to replace human manipulation, therefore risks of injury is reduced and efficiency is improved.

Research of motion planning is going through a flourishing period, due to development and popularity of *deep learning* (DL) and *reinforcement learning* (RL) that have better performance in coping with non-linear and complex problems. Many universities, tech giants, and research groups all over the world therefore attach much importance, time, and energy on developing new motion planning techniques by applying DL algorithms or integrating traditional motion planning algorithms with advanced *machine learning* (ML) algorithms. Autonomous vehicle is an example. Among tech giants, Google initiated their self-driving project named Waymo in 2016. In 2017, Tesla pledges a fully self-driving capable vehicle. Autonomous car from Baidu has successfully been tested in highways near Beijing in 2017, and man-manipulated buses from Huawei have already been replaced by automated buses in some specific areas of Shenzhen. Other companies in traditional vehicle manufacturing, like Audi and Toyota, also have their own experimental autonomous vehicles. Among research institutes and universities, Navlab (navigation lab) in Carnegie Mellon, Oxford University and MIT are leading research institutes. Up to 2020, European countries like Belgium, France, Italy, and UK are planning to operate transport systems for autonomous vehicles. Twenty-nine US states have passed laws in permitting autonomous vehicles. Autonomous vehicle is therefore expected to widely spread in near future with improvement of *traffic laws*.

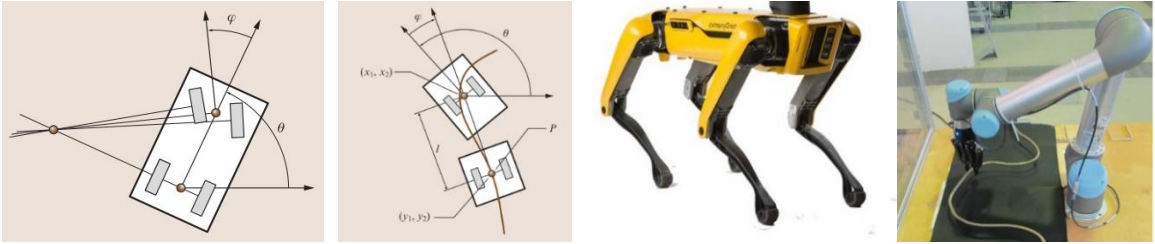

Figure. 1 Three types of robotic platform. First and second figures represent wheel-based chassis [37]. The first figure represents a car-like chassis, while the second figure represents a differential-wheel chassis. Third and fourth figures represent four-leg dog "SpotMini" from Boston Dynamic" and robotic arm [54].

**Motion planning and robotic platform:** Robotics use motion planning algorithms to plan their trajectories both at global and local level. Human-like and dog-like robotics from Boston Dynamic and autonomous robotic car from MIT [2] are good examples. All of them leverage motion planning algorithms to enable robotics to freely walk in complex and dynamic scenarios both indoor and outdoor. *Chassis of robotics* have two types of wheels, including *car-like wheel* and *ddifferential wheel* (Figure 1).

In robotics with car-like wheels, front two wheels are utilized for steering, while rear two wheels is used for forwarding. The car-like chassis has two servos. Front two wheels share a same servo and it means these two wheels can steer with a same steering angle or range $\varphi$ (Fig. 1). Rear two wheels share another servo to control the speed of robotics. Robotics using differential wheel, however, is completely different with car-like robot in functions of servo. The chassis with differential wheels generally has two servos, and each wheel is controlled by one servo for forwarding, and steering is realized by giving different speeds to each wheel. Steering range in car-like robotics is limited because two front wheels steer with a same angle $\varphi$. The car-like wheel is therefore suitable to be used in high-speed outdoor scenarios because of stability. Robotics with differential wheels, however, can steer in an angle of $2\pi$, and it means robotics can change their pose arbitrarily without moving forward. Robotics with differential wheels is also sensitive to the speed difference of its two front wheels, and it means it is flexible to move in low-speed indoor scenarios but very dangerous to move in high-speed situations if something wrong in speed control of two front wheels, because little speed changes of front two wheels in differential chassis can be exaggerated and accident follows.

It is popular to use *legs* in the chassis of robotics in recent years. Typical examples are human-like and animal-like (dog-like, Fig. 1) robotics from Boston Dynamic. Robotic *arms* (Fig. 1) are also a popular platform to deploy motion planning algorithms. In summary, wheels, arms, and legs are choices of chassis to implement motion planning algorithms which are widely used in academic and industrial scenarios including commercial autonomous driving, service robot, surgery robot and industrial arms.

**Architecture of robotics:** Basic architecture of automated robotics can be divided into modules that include data collection, environment perception and understanding, decision making and decision execution (Fig. 2). First, data are collected from sensors like *light detection and ranging* (liDAR) and camera. Data are processed and interpreted by advanced algorithms like motion planning, path planning, lane detection and tracking algorithms in environment perception and understanding processes. Then, decisional messages are generated according to outputs of mentioned algorithms. Finally, these messages are parsed from digital format into analog format that can be recognized and executed by hardware.

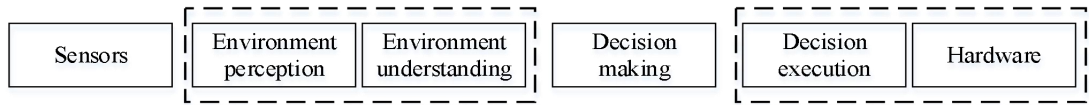

Figure 2. Basic architecture of robotics.

**Motion planning and path planning:** Performance of motion planning directly decides whether task of robotics is successful or not, therefore requiring much more attention than other functional modules in robotics. *Motion planning* is extension of *path planning*. They are almost the same term, but few differences exist. For example, path planning aims at finding the path between origin and destination in *workspace* by strategies like shortest distance or shortest time (Fig. 3), therefore path is planned from topological level. Motion planning, however, aims at generating interactive trajectories in workspace when robots interact with dynamic environment, therefore motion planning needs to consider kinetics features, velocities and poses of robots and dynamic objects nearby (Fig. 3). In short, motion planning considers short-term optimal or suboptimal strategies where robots interact with the environment to achieve long-term optimal motion planning strategy. Denote that workspace is an area that an algorithm works or the task exists.

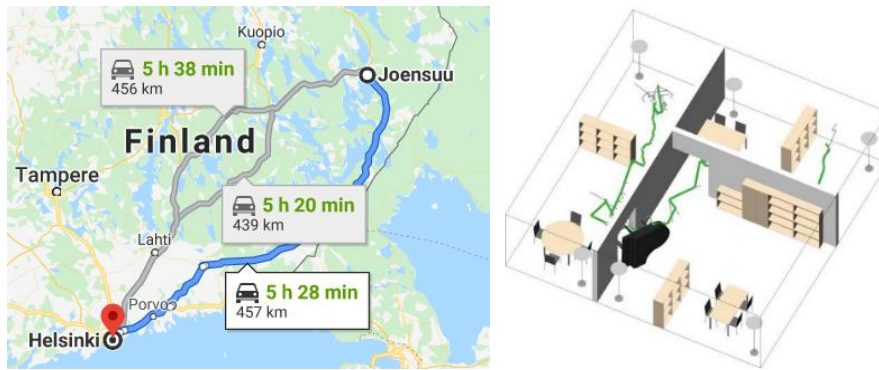

Figure 3. Path planning and motion planning. The left figure represents a path based on shortest distance and time, and path is generated from topological level. The right figure represents famous piano mover's problem that not only consider planning a path from topological level, but also consider kinetics features, speeds and poses of the piano.

**Classification of planning algorithms:** We divide motion planning algorithms into two categories: *traditional algorithms* and *ML-based* algorithms according to their principles and the era they were invented. Traditional algorithms are composed by three groups including *graph search algorithms*, *sampling-based algorithms* and *interpolating curve algorithms*. ML based planning algorithms are based on ML approaches that include *supervised learning* (e.g. *support vector* *machine* (SVM) [53]), *optimal value RL* and *policy gradient RL*. Categories of planning algorithms are summarized in Fig. 4.

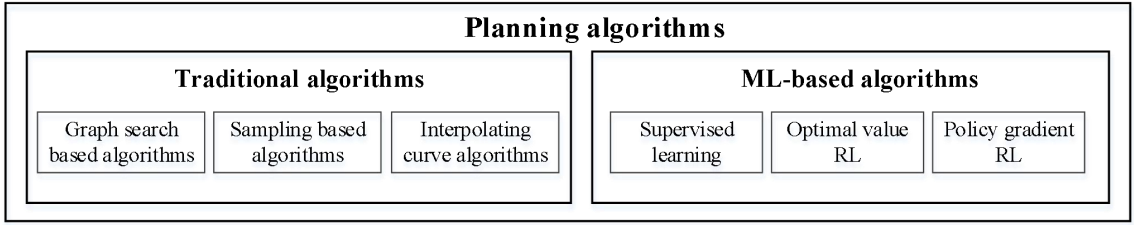

Figure 4. Classification of planning algorithms.

**Development of ML-based algorithms:** Development of ML-based motion planning algorithms is shown in Fig. 5. Researchers use supervised learning, like SVM, to implement simple motion planning at an earlier stage, but its performance is poor because SVM is short-sighted (one-step prediction) and it requires well-prepared vector inputs that cannot fully represent features of image-based dataset. Significant improvement to extract high-level features from images were made after invention of *convolutional neural network* (CNN) [34]. CNN is widely used in many image-related tasks including motion planning, but it cannot cope with complex time-sequential motion planning problems. These better suit *Markov chain* [60] and *long short-term memory* (LSTM) [4]. Many researchers combine neural networks with LSTM or algorithms that are based on Markov chain (e.g. *Q learning* [28]) to implement time-sequential motion planning tasks. However, their efficiency is limited. A breakthrough was made when Google DeepMind introduced nature *deep Q-learning network* (DQN) [38-39], in which *reply buffer* is to reuse old data to improve efficiency. Performance in robustness, however, is limited because of noise that impacts estimation of $Q$ value. *Double DQN* [40][42] and *dueling DQN* [5] are therefore invented to cope with noise in DQN. Double DQN utilizes another network to evaluate the estimation of $Q$ value in DQN to reduce noise, while advantage value ($A$ value) is utilized in dueling DQN to obtain better $Q$ value, and noise is mostly reduced. The $Q$ learning, DQN, double DQN and dueling DQN are all based on optimal values ($Q$ value and $A$ value) to select time-sequential actions, therefore these algorithms are called *optimal value algorithms*. Implementation of optimal value algorithms, however, is computationally expensive.

Optimal value algorithms are latter replaced by *policy gradient method* [43], in which *gradient approach* [59] is directly utilized to upgrade policy that is used to generate optimal actions. Policy gradient method is more stable in network convergence, but it lacks efficiency in speed of network convergence. *Actor-critic algorithm* [6][44] improves speed of convergence by actor-critic architecture. However, improvement in convergence speed is achieved by sacrificing the stability of convergence, and it is hard to converge in earlier-stage training. *Asynchronous advantage actor-critic* (A3C) [33][45], *advantage actor-critic* (A2C) [29][36], *trust region policy optimization* (TRPO) [69] and *proximal policy optimization* (PPO) [70] algorithms are then invented to cope with this shortcoming. *Multi-thread technique* [45] is utilized in A3C and A2C to accelerate the speed of convergence, while TRPO and PPO improve the policy of actor-critic algorithm by introducing *trust region constraint* in TRPO, and "surrogate" and adaptive penalty in PPO to improve speed and stability of convergence. Data, however, is dropped after training, and new data must therefore be collected to train the network until convergence of network.

*Off-policy gradient algorithms* including *deterministic policy gradient* (DPG) [47] and *deep DPG* (DDPG) [46][67] are invented to reuse data by replay buffer of DQN. DDPG fuses the actor-critic architecture and *deterministic policy* to enhance the performance in network convergence. In summary, supervised learning, optimal value RL, and policy gradient RL are typical ML algorithms in motion planning.

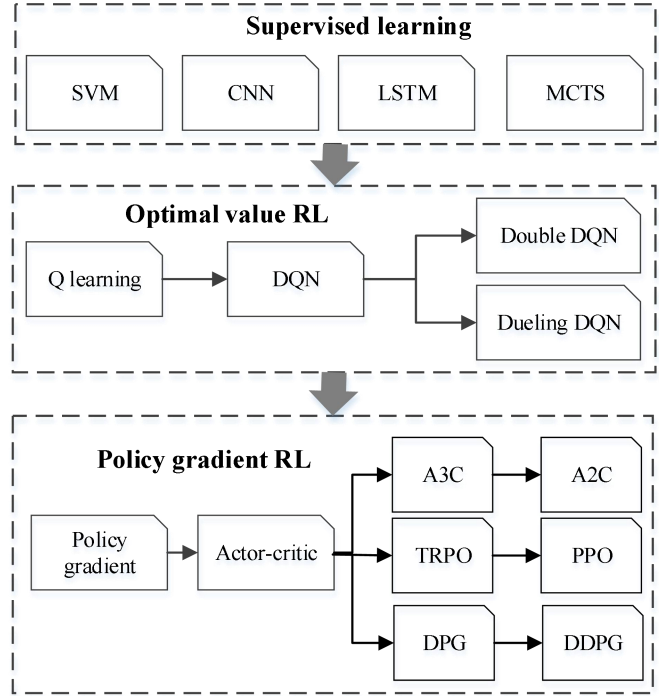

Figure 5. Development of ML based motion planning algorithms. ML-based motion planning algorithms evolve from supervised learning to optimal value RL and policy gradient RL. Supervised learning cannot address time-sequential planning problem but RL addresses it well. Optimal value RL suffers slow and unstable convergence speed but policy gradient RL performs better in convergence. Note that *Monte-carlo tree search* (MCTS) is a traditional RL algorithm but in this paper, we place it in group of supervised learning for convenient and clear comparisons.

In this paper, we investigate and analyze state-of-art ML based algorithms to provide researchers with a comprehensive and clear understanding about functions, structures, advantages, and disadvantages of planning algorithms. We also introduce new criteria to evaluate the performance of planning algorithms. Potential directions for making practical optimization in motion planning algorithms are discussed simultaneously. Contributions of this paper include (1) General survey of traditional planning algorithms; (2) Detailed survey of supervised learning, optimal value RL and policy gradient RL for robotic motion planning; (3) Analytical comparisons of these algorithms according to new evaluation criteria; (4) Analysis of future directions.

This paper is organized as follows: sections II, III, IV and V describe principles and applications of traditional planning algorithms, supervised learning, optimal value RL and policy gradient RL in robotic motion planning; section VI describes analytical comparisons of these algorithms, and criteria for performance evaluation; section VII analyzes future direction of robotic motion planning.

# II. Traditional planning algorithms

Traditional planning algorithms can be divided into three groups: *graph-search*, *sampling-based* and *interpolating curve* algorithms. They will be described in detail in the following sections.

**2.1 Graph-search algorithms**

Graph-search-based algorithms can be divided into *depth-first search*, *breadth-first search*, and *best-first search* [7]. The depth-first search algorithm builds a search tree as deep and fast as possible from origin to destination until a proper path is found. The breadth-first search algorithm shares similarities with the depth-first search algorithm by building a search tree. The search tree in the breadth-first search algorithm, however, is accomplished by extending the tree as broad and quick as possible until a proper path is found. The best-first search algorithm adds a numerical criterion (value or cost) to each node and edge in the search tree. According to that, the search process is guided by calculation of values in the search tree to decide: (1) whether search tree should be expanded; (2) which branch in the search tree should be extended. The process of building search trees repeats until a proper path is found. Graph search algorithms are composed by many algorithms. The most popular are *Dijkstra's algorithm* [7] and *A\* algorithm* [8].

**Dijkstra's algorithm** is one of earliest optimal algorithms based on best-first search technique to find the shortest paths among nodes in a graph. Finding the shortest paths in a road network is a typical example. Steps of the Dijkstra algorithm (Fig. 6) are as follows: (1) convert the road network to a graph, and distances between nodes in the graph are expected to be found by exploration; (2) pick the unvisited node with the lowest distance from the source node; (3) calculate the distance from the picked node to each unvisited neighbor and update the distance of all neighbor nodes if the distance to the picked node is smaller than the previous distance; (4) mark the visited node when the calculation of distance to all neighbors is done. Previous steps repeat until the shortest distance between origin and destination is found. Dijkstra's algorithm can be divided into two versions: forward version and backward version. Calculation of overall cost in the backward version, called *cost-to-come*, is accomplished by estimating the minimum distance from selected node to destination, while estimation of overall cost in the forward version, called *cost-to-go*, is realized by estimating the minimum distance from selected node to the initial node. In most cases, nodes are expanded based on the *cost-to-go*.

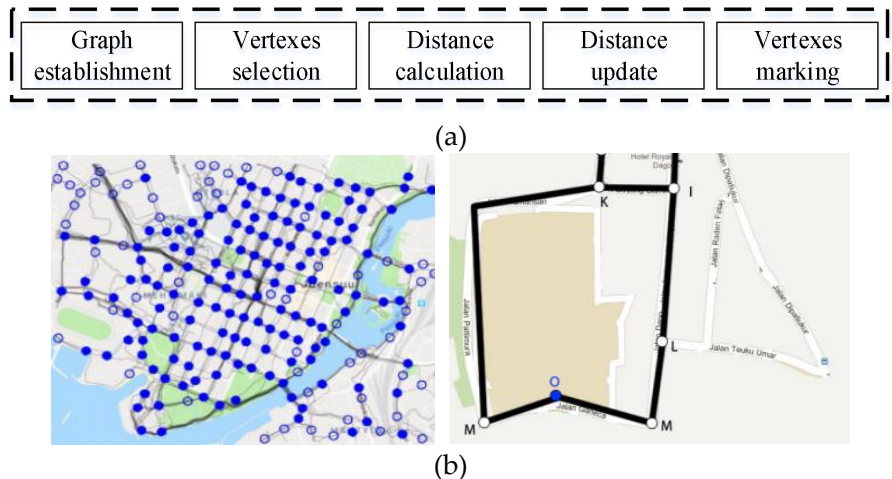

Figure 6. Steps of the Dijkstra algorithm (a) and road networks in web maps (b) [64-65]. Web maps are based on GPS data. Road network is mapped into the graph that is composed by nodes and edges, therefore graph search algorithms like A* and Dijkstra's algorithms can be used in these graphs.

**A\* algorithm** is based on the best-first search, and it utilizes heuristic function to find the shortest path by estimating the overall cost. The algorithm is different from the Dijkstra's algorithm in the estimation of the path cost. The cost estimation of a node $i$ in a graph by A* is as follows: (1) estimate the distance between the initial node and node $i$; (2) find the nearest neighbor $j$ of the node $i$, and estimate the distance of nodes $j$ and $i$; (3) estimate the distance

between the node *j* and the goal node. The overall estimated cost is the sum of these three factors:

$$C_i = c_{start,i} + min_j(d_{i,j} + d_{j,goal}). \tag{1}$$

where $C_i$ represents overall estimated cost of node *i*, $c_{start,i}$ the estimated cost from the origin to the node *i*, $d_{i,j}$ the estimated distance from the node *i* to its nearest node *j*, and $d_{j,goal}$ the estimated distance from the node *j* to the node of goal. A* algorithm has a long history in path planning in robotics. A common application of the A* algorithm is mobile rovers planning via an occupancy grid map (Fig. 7) using the Euclidean distance [9]. There are many variants of A* algorithm, like *dynamic A*ated* and *dynamic D** [10], *Field D** [11], *Theta** [12], *Anytime Repairing A** (ARA*) and *Anytime D** [13], *hybrid A** [14], and *AD** [15]. Other graph search algorithms have a difference with common robotic grid map. For example, the *state lattice algorithm* [16] uses one type of grid map with a specific shape (Fig. 7), while the grid in common robotic map is in a square-grid shape (Fig. 7).

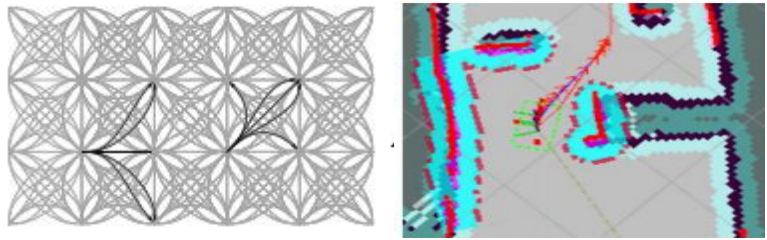

Figure 7. The left figure represents a specific grid map in the State Lattice algorithm, while the right figure represents a common square-grid (occupancy grid) map in the robot operating system (ROS).

## 2.2 Sampling-based algorithms

Sampling-based algorithms randomly sample a fixed workspace to generate sub-optimal paths. The *rapidly-exploring random tree* (RRT) and the *probabilistic roadmap method* (PRM) are two algorithms that are commonly utilized in motion planning. The RRT algorithm is more popular and widely used for commercial and industrial purposes. It constructs a tree that attempts to explore the workspace rapidly and uniformly via a random search [17]. The RRT algorithm can consider non-holonomic constraints, such as the maximum turning radius and momentum of the vehicle [18]. The example of trajectories generated by RRT is shown in Fig. 8. The PRM algorithm [20] is normally used in a static scenario. It is divided into two phases: *learning phase* and *query phase*. In the learning phase, a collision-free probabilistic roadmap is constructed and stored as a graph. In query phase, a path that connects original and targeted nodes, is searched from the probabilistic roadmap. An example of trajectory generated by PRM is shown in Fig. 8.

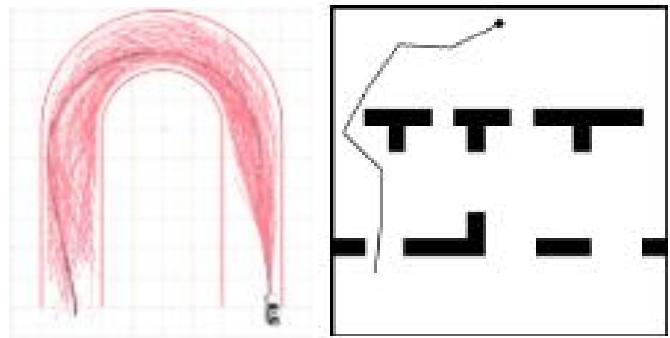

Figure 8. Trajectories of RRT and PRM. The left figure represents trajectories of RRT algorithm [19], and the right figure represents the trajectory of PRM algorithm [20].

## 2.3 Interpolating curve algorithms

*Interpolating curve algorithm* is defined as a process that constructs or inserts a set of mathematical rules to draw trajectories. The interpolating curve algorithm is based on techniques (e.g. *computer aided geometric design* (CAGD)) to draw a smooth path. Mathematical rules are used for path smoothing and curve generation. Typical path smoothing and curve generation rules include *line and circle* [21], *clothoid curves* [22], *polynomial curves* [23], *Bezier curves* [24] and *spline curves* [25]. Examples of trajectories are shown in Fig. 9.

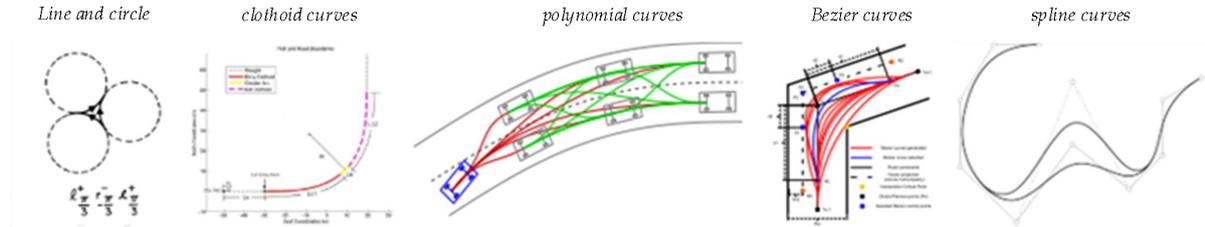

Figure 9. Interpolating curve algorithms generated by mathematical rules [21-25].

# III. Supervised learning

Here we present basic principle of 4 pervasive supervised learning algorithms for motion planning. These include SVM, LSTM, MCTS and CNN.

**SVM** [53] is a well-known supervised learning algorithm for classification. The basic principle of SVM is about drawing an optimal separating hyperplane between inputted data by training a maximum margin classifier [53]. Inputted data is in the form of vector that is mapped into high-dimensional space where classified vectors are obtained by trained classifier. SVM is used in 2-class classification that cannot suit real-world task, but its variant *multiclass SVM* (MSVM) [71] works.

**LSTM** [72][4] is a variant of *recurrent neural network* (RNN). LSTM can remember inputted data (vectors) in its cells. Because of limited capacity of cell in storage, a part of data will be dropped when cells are updated with past and new data, and then a part of data will be remembered and transferred to next time step. These functions in cells are achieved by neural network as the description in Fig. 10. In robotic motion planning, robots' features and labels in each time step are fed into neural networks in cells for training, therefore decisions for motion planning are made by performing trained network.

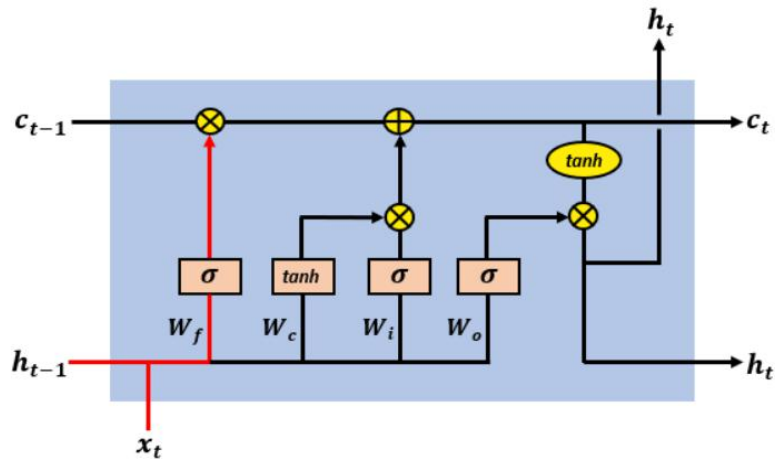

Figure 10. Cells of LSTM that are implemented using neural network [73]. $c_t$ denotes cell's state in time step $t$. $h_t$ denotes the output that will be transferred to the next state as its input,

therefore format of input is the vector $[h_{t-1}, x_t]$. Cell states are controlled and updated by 3 gates (forget gate, input gate and output gate) that are implemented using neural networks with weights $W_f$, $W_c + W_i$, and $W_o$ respectively.

**MCTS** is the combination of Monte-carlo method [75] and search tree [76]. MCTS is widely used in games (e.g. Go and chess) for motion prediction [74][3]. Mechanism of MCTS is composed by 4 processes that include selection, expansion, simulation, and backpropagation as Fig. 11. In robotic motion planning, node of MCTS represents possible state of robot, and stores state value of robot in each step. First, selection is made to choose some possible nodes in the tree based on known state value. Second, tree expands to unknown state by tree policy (e.g. random search). Third, simulation of expansion is made on new-expanded node by default policy (e.g. random search) until terminal state of robot and reward *R* is obtained. Finally, backpropagation is made from new-expanded node to root node, and state values in these nodes are updated according received reward. These four processes repeat until convergence of state values in the tree, therefore robot can plan its motion according to state values in the tree. MCTS fits discrete-action tasks (e.g. AlphaGo [74]), and it also fits time-sequential tasks like autonomous driving.

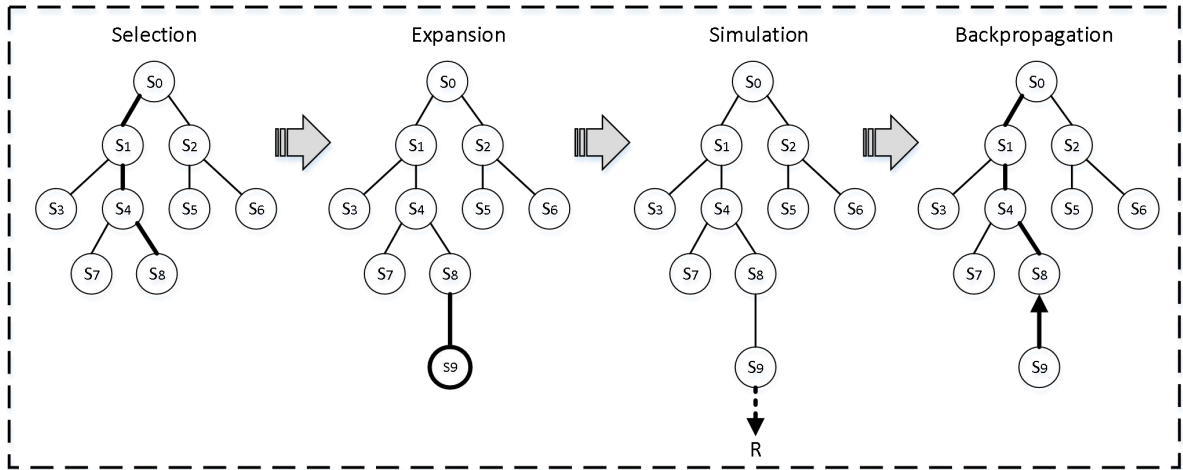

Figure 11. 4 processes of MCTS. These processes repeat until the convergence of state values in the tree.

**CNN** [34] has become a research focus of ML after *LeNet5* [34] was introduced and successfully applied into handwritten digits recognition. CNN is one of the essential types of neural network because it is good at extracting high-level features from high-dimensional high-resolution images by convolutional layers. CNN makes the robot avoid obstacles and plans motions of robot according to human experience by models trained in *forward propagation* and *back propagation* process, especially the back propagation. In the back propagation, a model with a weight matrix/vector $\theta$ is updated to record features of obstacles. Note that $\theta = \{w_i, b_i\}_i^L$ where *w* and *b* represent weight and bias, and *i* represents the serial number of *w-b* pairs. *L* represents the length of weight.

Training steps of CNN are shown as Fig. 12. Images of objects (obstacles) are used as inputs of CNN. Outputs are *probability distributions* obtained by *Softmax function* [58]. *Loss value* $Loss_{CE}$ is *cross-entropy* (CE) and that is obtained by

$$Loss_{CE} = -\sum_i p_i \cdot \log q_i \qquad (2)$$

where *p* denotes probability distributions of output (observed real value), *q* represents probability distributions of expectation ($p, q \in (0,1)$), and *i* represents the serial number of each batch of images in training. The loss function measures the difference (distance) of observed real value *p* and expected value *q*. *Mean-square error* (MSE) is an alternative of CE and MSE is defined by $Loss_{MSE} = \sum_i (p_i - q_i)^2$ where $p_i$ represents observed values while $q_i$ represents predicted values or expectation. The weight is updated in optimizer by minimizing the loss value using *gradient descent approach* [59] therefore new weight $w_i^{new}$ is obtained by

$$w_i^{new} = w_i - \eta \cdot \frac{\partial Loss}{\partial w_i} \quad (3)$$

where *w* represents the weight, *η* represents a learning rate (η ∈ (0,1)) and *i* represents the serial number of each batch of images in training. Improved variants of CNN is also widely used in motion planning, e.g. *residue networks* [35][49].

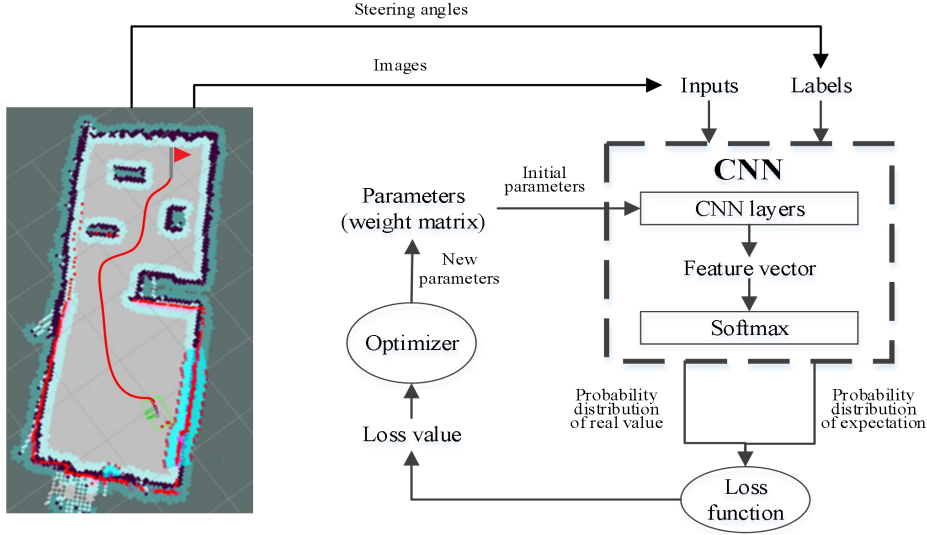

Figure 12. Training steps of CNN. The trajectory is planned by human in data collection in which steering angles of robotics are recorded as labels of training data. Robotics learn behavior strategies in training and move along the planned trajectory in the test. The Softmax function maps values of feature to probabilities between 0 and 1. The optimizer represents gradient descent approach, e.g. *stochastic gradient descent* (SGD) [59].

# IV. Optimal value RL

Here we first introduce basic concepts in RL, and then introduce principles of Q learning, nature DQN, double DQN and dueling DQN.

Supervised learning algorithms like CNN is competent only in static obstacle avoidance by one-step prediction, therefore it cannot cope with time-sequential obstacle avoidance. RL algorithms, e.g. optimal value RL, fit time-sequential tasks. Typical examples of these algorithms include Q learning, nature DQN, double DQN and dueling DQN. Motion planning is realized by attaching destination and safe paths with big *reward* (numerical value), while obstacles are attached with *penalties* (negative reward). Optimal path is found according to total rewards from initial place to destination. To better understand optimal value RL, it is necessary to recall several fundamental concepts: *Markov chain*, *Markov decision process* (MDP), *model-based dynamic programming*, *model-free RL*, *Monte-Carlo method* (MC), *temporal difference method* (TD), and *State-action-reward-state-action* (SARSA). MDP is based on Markov chain [60], and it can be divided into two categories: model-based dynamic programming and model-free RL. Mode-free RL can be divided into MC and TD that includes SARSA and Q learning algorithms. Relationship of these concepts is shown in Fig. 13.

**Markov chain:** Variable set $X = \{X_n : n > 0\}$ is called Markov chain [60] if *X* meets

$$p(X_{t+1}|X_t,...,X_1) = p(X_{t+1}|X_t). \tag{4}$$

This means the occurrence of event $X_{t+1}$ depends only on event $X_t$ and has no correlation to any earlier events.

**Markov decision process:** MDP [60] is a sequential decision process based on Markov Chain. This means the state and action of the next step depend only on the state and action of the current step. MDP is described as a tuple $<S,A,P,R>$. $S$ represents state and here refers to states of robot and obstacles. $A$ represents an action taken by robot. State $S$ transits into another state under a state-transition probability $P$ and a reward $R$ from the environment is obtained. Principle of MDP is shown in Fig. 13. First, the robot in state $s$ interacts with the environment and generate an action based on policy $\pi(s): s \to a$. Robot then obtains the reward $r$ from the environment, and state transits into the next state $s'$. The reach of next state $s'$ marks the end of one loop and the start of the next loop.

**Model-free RL and model-based dynamic programming:** Problems in MDP can be solved using *model-based dynamic programming* and *model-free RL* methods. The model-based dynamic programming is used in a known environment, while the model-free RL is utilized to solve problems in an unknown environment.

**Temporal difference and Monte Carlo methods:** The model-free RL includes MC and TD. A sequence of actions is called an *episode*. Given an episode $<S_1, A_1, R_2, S_2, A_2, R_3, ..., S_t, A_t, R_{t+1}, ..., S_T>$, the state value $V(s)$ in the time step $t$ is defined as the expectation of overall rewards $G_t$ by

$$V(s) = \mathbb{E}[G_t = R_{t+1} + \gamma R_{t+1} + ... + \gamma^{T-1} R_T | S_t = s] \tag{5}$$

where $\gamma$ represent a discount factor ($\gamma \in [0,1]$). MC uses $G_t - V(s)$ to update its state value $V_{MC}(s)$ by

$$V_{MC}(s) \leftarrow V(s) + \alpha(G_t - V(s)) \tag{6}$$

where "$\leftarrow$" represents the update process in which new value will replace previous value. $\alpha$ is a discount factor. TD uses $R_{t+1} + \gamma V(s_{t+1}) - V(s)$ to update its state value $V_{TD}(s)$ by

$$V_{TD}(s) \leftarrow V(s) + \alpha[R_{t+1} + \gamma V(s_{t+1}) - V(s)] \tag{7}$$

where $\alpha$ is a learning rate, $R_{t+1} + \gamma V(s_{t+1})$ is *TD target* in which the estimated state value $V(s_{t+1})$ is obtained by *bootstrapping* method [56]. This means MC updates its state value after the termination of an episode, while TD update its state value in every steps. TD method is therefore efficient than MC in state value update.

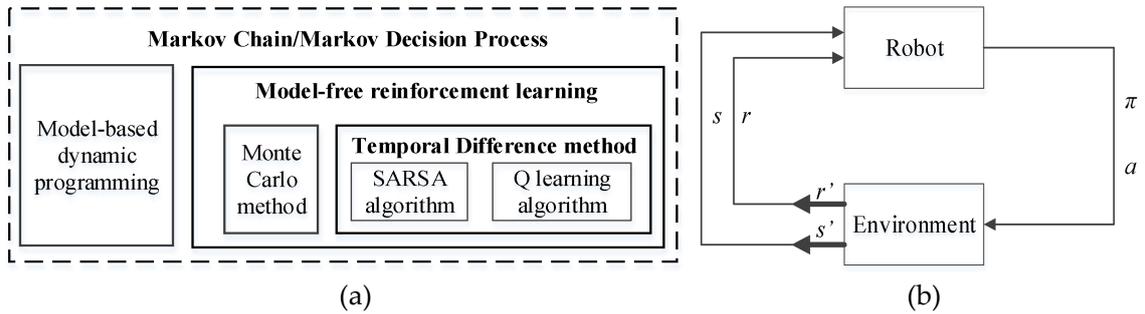

(a)  (b)

Figure 13. (a) represents the relationship of basic concepts. (b) represents the principle of MDP.

## 4.1 Q learning

TD includes SARSA [55] and Q learning [28][66]. Given an episode $<S_1, A_1, R_2, S_2, A_2, R_3, ..., S_t, A_t, R_{t+1}, ..., S_T>$, SARSA and $Q$ learning use the *ε-greedy* method [57] to select an action $A_t$ at time step $t$. There are two differences between SARSA and $Q$ learning: (1) SARSA uses *ε-greedy* again to select an estimated action value $Q(S_{t+1}, A_{t+1})$ at time step $t$+1 to update its action value by

$$Q_{SARSA}(S_t,A_t) \leftarrow Q(S_t,A_t) + \alpha(R_{t+1} + \gamma Q(S_{t+1},A_{t+1}) - Q(S_t,A_t)), \qquad (8)$$

while $Q$ learning directly uses maximum estimated action value $\max_Q Q(S_{t+1},A_{t+1})$ at time step $t$+1 to update its action value by

$$Q_{QL}(S_t,A_t) \leftarrow Q(S_t,A_t) + \alpha(R_{t+1} + \gamma \max_{A_{t+1}} Q(S_{t+1},A_{t+1}) - Q(S_t,A_t)); \qquad (9)$$

(2) SARSA adopts selected action $A_{t+1}$ directly to update its next action value, but $Q$ learning algorithm use *ε-greedy* to select a new action to update its next action value.

SARSA uses *ε-greedy* method to sample all potential action value of next step and selects a "safe" action eventually, while $Q$ learning pays attention to the maximum estimated action value of the next step and selects optimal actions eventually. Steps of SARSA is shown in **Algorithm 1** [66], while $Q$ learning algorithm as **Algorithm 2** [66] and Fig. 14. Implementations of robotic motion planning by Q learning are as [28][30][50].

---

**Algorithm 1:** SARSA

Initialize $Q(s,a)$ arbitrarily
**Repeat** (for each episode):
  Initialize $s$
  Choose $a$ from $s$ using policy derived from $Q$ (e.g. *ε-greedy*)
  **Repeat** (for each step of episode):
    Take action $a$, observe $r$, $s'$
    Choose $a'$ from $s'$ using policy derived from $Q$ (e.g. *ε-greedy*)
    $Q(s,a) \leftarrow Q(s,a)+\alpha[r+\gamma Q(s',a')-Q(s,a)]$
    $s \leftarrow s'; a \leftarrow a'$;
  **until** $s$ is terminal

---

**Algorithm 2:** Q-learning

Initialize $Q(s,a)$ arbitrarily
**Repeat** (for each episode):
  Initialize $s$
  **Repeat** (for each step of episode):
    Choose $a$ from $s$ using policy derived from $Q$ (e.g. *ε-greedy*)
    Take action $a$, observe $r$, $s'$
    $Q(s,a) \leftarrow Q(s,a)+\alpha[r+\gamma \max_{a'} Q(s',a')-Q(s,a)]$
    $s \leftarrow s'$;
  **until** $s$ is terminal

---

Note that $s'$ and $a'$ denote $S_{t+1}$ and $A_{t+1}$ respectively.

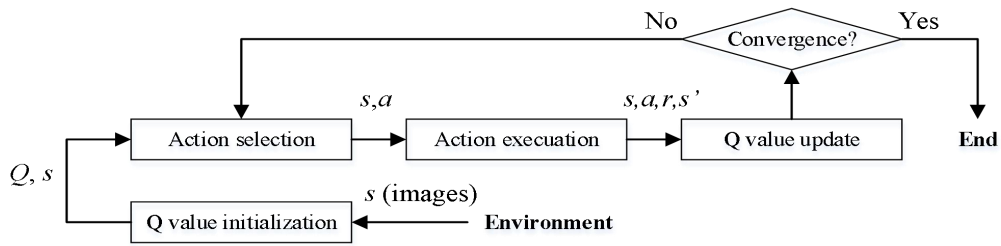

Figure 14. Steps of $Q$ learning algorithm. Input of Q learning is in the vector format normally. Q value is obtained via Q value table or network as approximator. Extra preprocessing is needed to extract features from image if input is in image format.

## 4.2 Nature deep Q-learning network

DQN [38] is a combination of *Q* leaning and deep neural network (e.g. CNN). DQN uses CNN to approximate Q values by its weight $\theta$. Hence, *Q* table in Q learning changes to *Q* value network that can be converged in a faster speed for complex motion planning. DQN became a research focus when it was invented by Google DeepMind [38][39], and performance of DQN approximates or even surpasses the performance of human being in Atari games (e.g. Pac-man and Enduro in Fig. 15) and real-world motion planning tasks [31][51]. DQN utilizes CNN to approximate *Q* values (Fig. 16) by

$$Q^*(s,a) \approx Q(s,a;\theta). \tag{10}$$

In contrast with the *Q* learning, DQN features 3 components: CNN, *replay buffer* [41] and *targeted network*. CNN extracts feature from images as its inputs. Outputs can be *Q* value of current state *Q*(*s*,*a*) and *Q* value of next state *Q*(*s'*,*a'*), therefore experiences <*s*,*a*,*r*,*s'*> are obtained and temporarily stored in replay buffer. It is followed by training DQN using experiences in the replay buffer. In this process, a targeted network is leveraged to minimize the loss value by

$$Loss = (r + \gamma \max_{a'} Q(s',a';\theta') - Q(s,a;\theta))^2. \tag{11}$$

Loss value measures the distance between expected value and real value. In DQN, expected value is (*r*+$\gamma$max*Q*(*s'*,*a'*;$\theta'$)) that is similar to labels in supervised learning, while *Q*(*s*,*a*;$\theta$) is the observed real value. weights of targeted network and *Q* value network share a same weight $\theta$. The difference is that weight of *Q* value network $\theta$ is updated in each step, while weight of targeted network $\theta'$ is updated in a long period of time. Hence, $\theta$ is updated frequently and $\theta'$ is more stable. It is necessary to keep targeted network stable, otherwise *Q* value network will be hard to converge. Detailed steps of DQN are shown as **Algorithm 3** [38] and Fig. 17.

---

**Algorithm 3:** Deep Q-learning with experience replay
---
Initialize replay memory $\mathcal{D}$ to capacity $\mathcal{N}$
Initialize action-value function *Q* with random weights
**for** episode = 1, *M* **do**
  Initialize sequence $s_1 = \{x_1\}$ and preprocessed sequenced $\emptyset_1 = \emptyset(s_1)$
  **for** *t* = 1, *T* **do**
    With probability $\epsilon$ select a random action $a_t$
    otherwise select $a_t = \max_a Q^*(\emptyset(s_t),a;\theta)$
    Execute action $a_t$ in emulator and observe reward $r_t$ and image $x_{t+1}$
    Set $s_{t+1} = s_t, a_t, x_{t+1}$ and preprocess $\emptyset_{t+1} = \emptyset(s_{t+1})$
    Store transition ($\emptyset_t, a_t, r_t, \emptyset_{t+1}$) in $\mathcal{D}$
    Sample random minibatch of transitions ($\emptyset_j, a_j, r_j, \emptyset_{j+1}$) from $\mathcal{D}$
    Set $y_j = \begin{cases} r_j & \text{for terninal } \emptyset_{j+1} \\ r_j + \gamma \max_{a'} Q(\emptyset_{j+1}, a';\theta) & \text{for nonternimal } \emptyset_{j+1} \end{cases}$
    Perform a gradient descent step on $(y_j - Q(\emptyset_j, a_j;\theta))^2$
  **end for**
**end for**

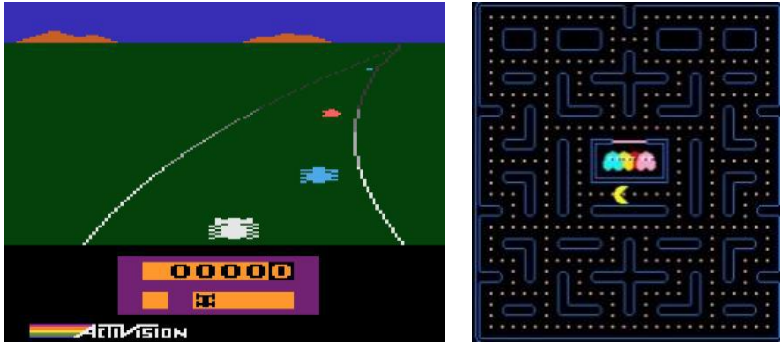

Figure 15. Two examples of motion planning in early-stage arcade games: Enduro (left) and Pac-man (right).

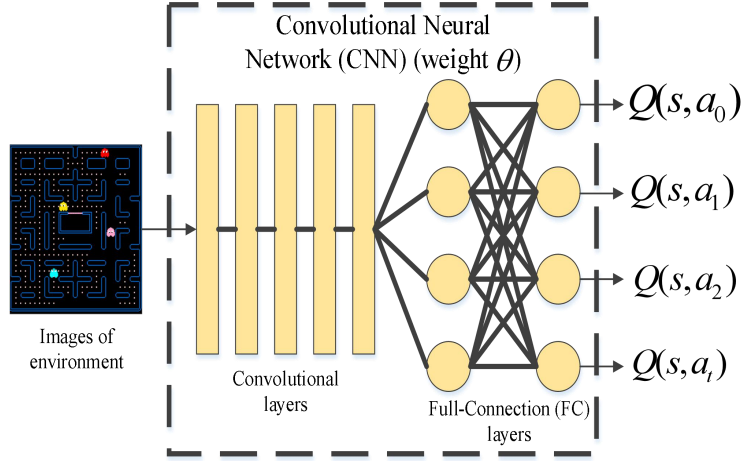

Figure 16. $Q(s,a_0)$, $Q(s,a_1)$, $Q(s,a_2)$ and $Q(s,a_t)$ denote $Q$ values of all potential actions.

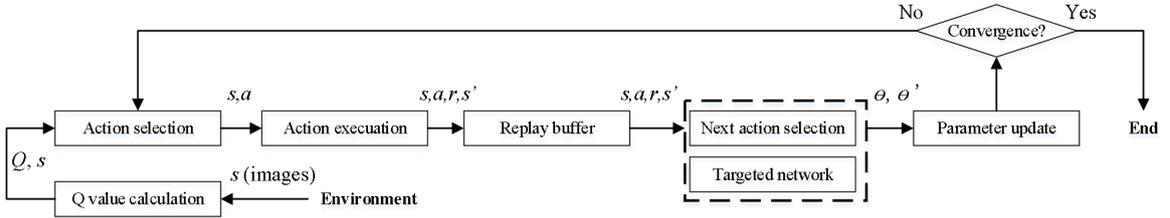

Figure 17. steps of DQN algorithm.

## 2.3 Double deep Q-learning network

Noise in DQN leads to bias and false selection of next action $a'$ follows, therefore leading to over-estimation of next action value $Q(s',a';\theta')$. To reduce the over-estimation caused by noise, researchers invented the *double DQN* [40] in which another independent targeted network with weight $\theta^-$ is introduced to evaluate the selected action $a'$. Hence, equation of targeted network therefore changes from $y^{DQN} = r + \gamma \max Q(s',a';\theta')$ to

$$y^{doubleDQN} = r + \gamma Q(s', \arg max_{a'} Q(s',a';\theta');\theta^-) \qquad (12)$$

Steps of double DQN are the same with DQN. Examples of application are [26][42][48] in which double DQN is used in games and *robotic operation system* (ROS).

## 2.4 Dueling deep Q-learning network

The state value $V^\pi(s)$ measures "how good the robot is" in the state $s$ where $\pi$ denotes policy $\pi: s \to a$, while the action value $Q^\pi(s,a)$ denotes "how good the robot is" **after** robot takes action *a* in state *s* using policy π. *Advantage value* (*A* value) denotes the difference of $Q^\pi(s,a)$ and $V^\pi(s)$ by

$$A(s,a) = Q(s,a) - V(s,a), \tag{13}$$

therefore *A* value measures "how good the action *a* is" in state *s* **if** robot takes action *a*. In neural network case (Fig. 18), weights α, β, θ are added, therefore

$$Q(s,a;\theta,\alpha,\beta) = V(s;\theta,\beta) + A(s,a;\theta,\alpha) \tag{14}$$

where θ is the weight of neural network and it is the shared weight of *Q*, *V* and *A* values. Here α denotes the weight of *A* value, and β the weight of *V* value. $V(s;\theta,\beta)$ is a scalar, and $A(s,a;\theta,\alpha)$ is a vector. There are however too many *V-A* value pairs if *Q* value is simply divided into two components, and only one *V-A* pairs are qualified. Thus, it is necessary to constrain the *V* value or *A* value to obtain a fixed *V-A* pair. According to relationship of $Q^\pi(s,a)$ and $V^\pi(s)$ where $V^\pi(s) = \mathbb{E}_{a \sim \pi(s)}[Q^\pi(s,a)]$, the expectation value of *A* is

$$\mathbb{E}_{a \sim \pi(s)}[A(s,a)] = 0. \tag{15}$$

Eq. 15 can be used as a rule to constrain *A* value for obtaining a stable *V-A* pair. Expectation of advantage value is obtained by using $A(s_t,a_t)$ to subtract mean *A* value that is obtained from all actions, therefore

$$\mathbb{E}[A(s_t,a_t)] = A(s_t,a_t) - \frac{1}{|\mathcal{A}|}\sum_{a_t^- \in \mathcal{A}} A(s_t,a_t^-) \tag{16}$$

where $\mathcal{A}$ represents *action space* in time step *t*, $|\mathcal{A}|$ number of actions, and $a_t^-$ one of actions in $\mathcal{A}$ in time step *t*. Expectation of *A* value keeps zero for $t \in [0,T]$, although the fluctuation of $A(s_t,a_t)$ in different action choices. Researchers use the expectation of *A* value to replace the current *A* value by

$$Q(s,a;\theta,\alpha,\beta) = V(s;\theta,\beta) + \left\{A(s,a;\theta,\alpha) - \frac{1}{|\mathcal{A}|}\sum_{a_t^- \in \mathcal{A}} A(s_t,a_t^-;\theta,\alpha)\right\}. \tag{17}$$

Thus, a stable *V-A* pair is obtained although original semantic definition of *A* value (Eq. 13) is changed [5]. In other words: (1) advantage constraint $\left\{A(s,a) - \frac{1}{|\mathcal{A}|}\sum_{a_t^- \in \mathcal{A}} A(s_t,a_t^-)\right\} = 0$ is used to find state advantage value; (2) action value is obtained by $Q(s,a;\theta,\alpha,\beta) = V(s;\theta,\beta)$ (this can be looked as a constraint) under $\left\{A(s,a) - \frac{1}{|\mathcal{A}|}\sum_{a_t^- \in \mathcal{A}} A(s_t,a_t^-)\right\} = 0$. Hence, 2 constraints lead to a better estimation of action value.

DQN obtained action value $Q(s,a)$ directly by using network to approximate action value. This process introduces over-estimation of action value. Dueling DQN obtains action value $Q(s,a)$ by constraining advantage value $A(s,a)$. Finally, 3 weights ($\theta,\alpha,\beta$) are obtained after training, and Q value network θ is with less bias but advantage value is better than action value to represent "how good the action is" (Fig. 19).

Further optimizations are distributional DQN [61], noise network [62], dueling double DQN [77] and rainbow model [63]. Distributional DQN is like the dueling DQN, as noise is reduced by optimizing the architecture of DQN. Noise network is about improving the ability in exploration by a more exquisite and smooth approach. Dueling double DQN and rainbow model are hybrid algorithms. Rainbow model fuses several high-performance algorithms as components that include double networks, replay buffer, dueling network, multi-step learning, distributional network, and noise network.

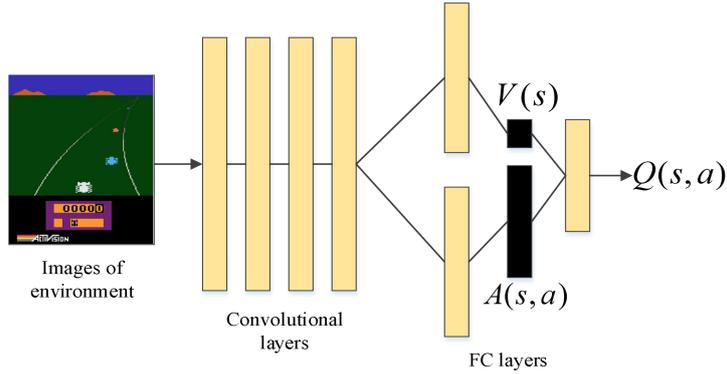

Figure 18. The architecture of dueling DQN, in which $Q$ value $Q(s,a)$ is decoupled into two parts, including $V$ value $V(s)$ and $A$ value $A(s,a)$.

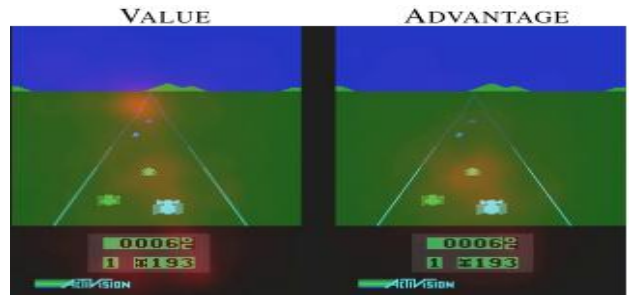

Figure 19. $Q(s,a)$ and $A(s,a)$ saliency maps (red-tinted overlay) on the Atari game (Enduro). $Q(s,a)$ learns to pay attention to the road, but pay less attention to obstacles in the front. $A(s,a)$ learns to pay much attention to dynamic obstacles in the front [5].

# V. Policy gradient RL

Here we first introduce policy gradient method and actor-critic algorithm, and then introduce their optimized variants: (1) A3C and A2C; (2) DPG and DDPG; (3) TROP and PPO.

Optimal value RL uses neural network to approximate optimal values to indirectly select action. This process is simplified as $a \leftarrow argmax_a R(s,a) + Q(s,a;\theta)$. Noise leads to over-estimation of $Q(s,a;\theta)$, therefore the selected actions are suboptimal, and network $\theta$ is hard to converge. Policy gradient algorithm uses neural network $\theta$ as policy $\pi_\theta : s \rightarrow a$ to directly select actions to avoid this problem. Brief steps of policy gradient algorithm are shown in Fig. 20.

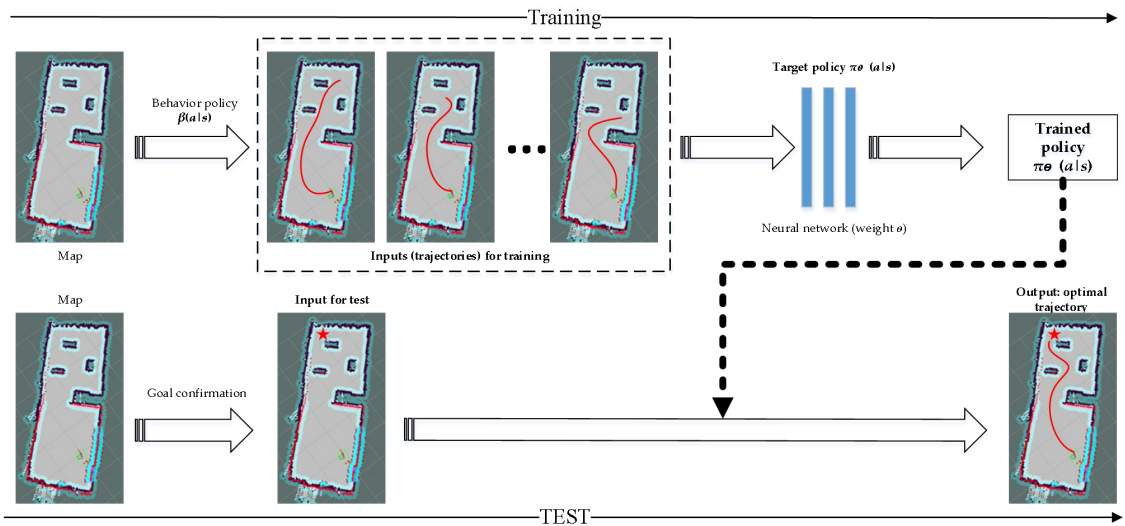

Figure 20. Training and test steps of policy gradient algorithms. In the training, trajectories are generated by behavior policy. Note that policy is divided to behavior policy and target policy. Behavior policy is about selecting action for training and behavior policy will not be updated, while target policy is also used to select actions but it will be updated in training. Policy refers to target policy normally. Robots learn trajectories via target policy (neural network as approximator) and trained policy is obtained. In the test, optimal actions are generated directly by trained policy until destination is reached.

### 5.1 Policy gradient method

Policy is a *probability distribution* $P\{a|s,\theta\}=\pi_\theta(a|s)=\pi(a|s,\theta)$ that is used to select action $a$ in state $s$, where weight $\theta$ is a parameter matrix that is used as an approximation of policy $\pi(a|s)$. *Policy gradient method* (PG) [43] seeks an optimal policy and uses it to find optimal actions. how to find this optimal policy? Given a trajectory $\tau=(s_1,a_1,...,s_T,a_T)$, the probability to output actions in $\tau$ is $\pi_\theta(\tau) = p(s_1) \prod_{t=2}^{T} \pi_\theta(a_t|s_t) p(s_t|s_{t-1},a_{t-1})$. The aim of the PG is to find optimal parameter $\theta^* = \arg\max_\theta \mathbb{E}_{\tau\sim\pi_\theta(\tau)}[R(\tau)]$ where trajectory reward $R(\tau) = \sum_{t=1}^{T} r(s_t,a_t)$) is overall reward in trajectory $\tau$. Objective of PG is defined as the expectation in trajectory $\tau$ by

$$J(\theta) = \mathbb{E}_{\tau\sim\pi_\theta(\tau)}[R(\tau)] = \int \pi_\theta(\tau) R(\tau) d\tau. \tag{18}$$

To find higher expectation of reward, gradient operation is used on objective to find the increment of network that leads to a better policy. Increment of network is the gradient value of objective, and that is

$$\nabla_\theta J(\theta) = \int \nabla_\theta \pi_\theta(\tau) R(\tau) d\tau = \int \pi_\theta(\tau) \nabla_\theta \log \pi_\theta(\tau) R(\tau) d\tau = \mathbb{E}_{\tau\sim\pi_\theta(\tau)}[\nabla_\theta \log \pi_\theta(\tau) R(\tau)]. \tag{19}$$

An example of PG is Monte-carlo reinforce [68]. Data $\tau$ for training are generated from simulation by *stochastic policy*. Previous objective and its gradient (Eq. 18-19) are replaced by

$$J(\theta) \approx \frac{1}{N} \sum_{i=1}^{N} \sum_{t=1}^{T} r(s_t^i, a_t^i) \tag{20}$$

$$\nabla_\theta J(\theta) \approx \frac{1}{N} \sum_{i=1}^{N} \left[\sum_{t=1}^{T} \nabla_\theta \log \pi_\theta(a_t^i, s_t^i)\right] \left[\sum_{t=1}^{T} r(s_t^i, a_t^i)\right] \tag{21}$$

where $N$ is the number of trajectories, $T$ the length of trajectory. A target policy $\pi_\theta$ is used to generate trajectory for training. For example, *Gaussian distribution function* is used as target policy to select actions by $a \sim N(\mu(s), \sigma^2)$. Network $f(s;\theta)$ is then used to approximate expectation of Gaussian distribution by $\mu(s) = f(s;\theta)$. It means $a \sim N(mean = f[s;\theta = \{w_i,b_i\}_i^L]), stdev = \sigma^2)$ and $\mu(s;\theta) = [mean, stdev]$ where $w$ and $b$ represent weight and bias of network, $L$ the number of $w$-$b$ pairs. Its objective is defined as $J(\theta) = \|f(s_t;w) - a_t\|_\Sigma^2$, therefore the objective gradient is

$$\nabla_\theta J(\theta) = -\frac{1}{2} \Sigma^{-1} (f(s_t) - a_t) \frac{df}{d\theta} \tag{22}$$

where $\frac{df}{d\theta}$ is obtained by backward-propagation. According to Eq. 21-22, the objective gradient is

$$\nabla_\theta J(\theta) \approx \frac{1}{N} \sum_{i=1}^{N} \left[\sum_{t=1}^{T} -\frac{1}{2} \Sigma^{-1} (f(s_t^i) - a_t^i) \frac{df}{d\theta}\right] \left[\sum_{t=1}^{T} r(s_t^i, a_t^i)\right]. \tag{23}$$

Once objective gradient is obtained, network is updated by gradient ascent method. That is

$$\theta \leftarrow \theta + \nabla_\theta J(\theta) \tag{24}$$

### 5.2 Actor-critic algorithm

The update of policy in PG is based on expectation of multi-step rewards in trajectory $\tau$ $\mathbb{E}_{\tau\sim\pi_\theta(\tau)}[R(\tau)]$. This leads to high *variance* that causes low speed in network convergence, but convergence stability is improved. *Actor-critic algorithm* (AC) [6][32][44] reduces the variance by one-step reward in TD-error $e$ for network update. TD-error is defined by

$$e = r_t + V(s_{t+1}) - V(s_t). \tag{25}$$

To enhance convergence speed, AC uses actor-critic architecture that includes *actor network (policy network)* and *critic network*. Critic network is used in TD-error to approximate state value by

$$e = r_t + V(s_{t+1};w) - V(s_t;w). \tag{26}$$

Objective of critic network is defined by

$$J(w) = e^2. \tag{27}$$

Objective gradient is therefore obtained by minimizing the mean-square error

$$\nabla_w J(w) = \nabla_w e^2. \tag{28}$$

Critic network is updated by *gradient ascent method* [59]. That is

$$w \leftarrow w + \beta \nabla_w J(w) \tag{29}$$

where $\beta$ represents learning rate. Objective of policy network is defined by

$$J(\theta) = \pi_\theta(a_t|s_t) \cdot e. \tag{30}$$

Hence, objective gradient of policy network is obtained by

$$\nabla_\theta J(\theta) = \nabla_\theta \log \pi_\theta(a_t|s_t) \cdot e \tag{31}$$

and policy network is updated by

$$\theta \leftarrow \theta + \alpha \nabla_\theta J(\theta) \tag{32}$$

where $\alpha$ is a learning rate of actor network. Detailed steps of the AC are as Fig. 21: (1) action $a_t$ at time step $t$ is selected by policy network $\theta$; (2) selected action is executed and reward is obtained. State transits into the next state $s_t$; (3) state value is obtained by critic network and TD error is obtained; (4) policy network is updated by minimizing objective of critic network; (5) critic network is updated according to objective gradient of critic network. This process repeats until the convergence of policy and critic networks.

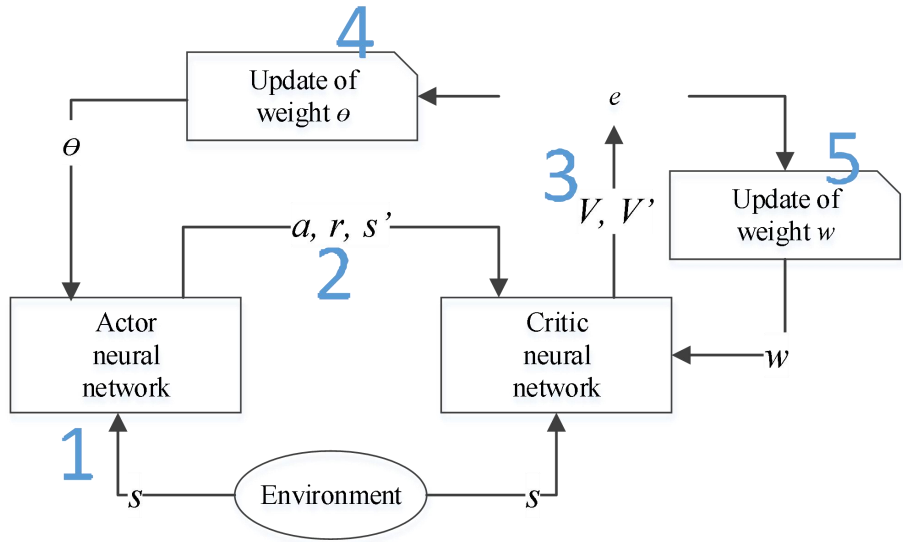

Figure 21. Training steps of AC.

**5.3 A3C and A2C**

**A3C:** in contrast to AC, the A3C [2] has three features: (1) multi-thread computing; (2) multi-step rewards; (3) policy entropy. Multi-thread computing means multiple interactions with the environment to collect data and update networks. Multi-step rewards are used in critic network, therefore the TD-error $e$ of A3C is obtained by

$$e = \sum_{i=t}^{T} \gamma^{i-t} r_i + V(s_{t+n}) - V(s_t) \tag{33}$$

therefore speed of convergence is improved. Here $\gamma$ is a discount factor, and $n$ is the number of steps. Data collection by policy $\pi(s_t;\theta)$ will cause *over-concentration*, because initial policy is with poor performance therefore actions are selected from small area of workspace. This causes poor quality of input, therefore convergence speed of network is poor. Policy entropy increases the ability of policy in action exploitation to reduce over-concentration. Objective gradient of A3C therefore changes to

$$\nabla_\theta J(\theta)_{A3C} = \nabla_\theta J(\theta)_{AC} + \beta \nabla_\theta H(\pi(s_t;\theta)) \tag{34}$$

where $\beta$ is a discount factor and $H(\pi(s_t;\theta))$ is the policy entropy.

**A2C:** A2C [29] is the alternative of A3C algorithm. Each thread in A3C algorithm can be utilized to collect data, train critic and policy networks, and send updated weights to global model. Each thread in A2C however can only be used to collect data. Weights in A2C are updated synchronously compared with the asynchronous update of A3C, and experiments demonstrate that synchronous update of weights is better than asynchronous way in weights update [36][45]. Their mechanisms in weight update are shown in Fig. 22.

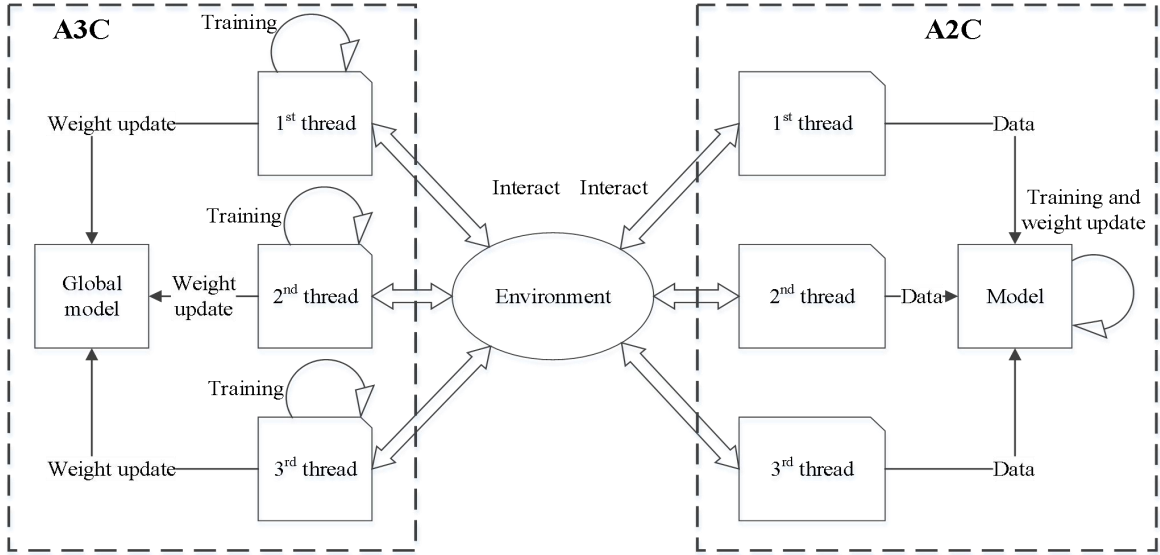

Figure 22. The weight update processes of the A3C and A2C.

## 5.4 DPG and DDPG

Here we first introduce prerequisites: *on-policy algorithm*, *off-policy algorithm*, *important sampling ratio*, *stochastic policy gradient algorithm*, and then introduce DPG and DDPG.

**Prerequisites:** in data generation and training processes, if behavior policy and target policy are the same policy $\pi_\theta$, these algorithms are called ***on-policy algorithm***. On-policy algorithm however may lead to low-quality data in data generation and a slow speed in network convergence. This problem can be reduced by using one policy (behavior policy) $\beta_\theta$ for data generation and another policy (target policy) $\pi_\theta$ for learning and making decision. Algorithms using different policies on data generation and learning are therefore called ***off-policy algorithms***. Although policies in off-policy algorithm are different, their relationship can still be measured by *transition probability* $\rho^\beta(s)$ that is the ***importance-sampling ratio*** and defined by

$$\rho^\beta(s) = \frac{\pi_\theta(a|s)}{\beta_\theta(a|s)} = \frac{\prod_{k=t}^T \pi(a_k|s_k;\theta)}{\prod_{k=t}^T \beta(a_k|s_k;\theta)}. \tag{35}$$

Importance-sampling ratio measures the similarity of two policies. These policies must be with large similarity in definition of important sampling. Particularly, behavior policy $\beta_\theta$ is the same as policy $\pi_\theta$ in on-policy algorithms. This means $\pi_\theta = \beta_\theta$ and $\rho^\beta(s)=\rho^\pi(s)=1$.

In on-policy policy gradient algorithm (e.g. PG), its objective is defined as

$$J(\theta) = \mathbb{E}_{\tau \sim \pi_\theta(\tau)}[R(\tau)] = \int_{\tau \sim \pi_\theta(\tau)} \pi_\theta(\tau) R(\tau) d\tau = \int_{s \sim S} \rho^\pi(s) \int_{a \sim A} \pi_\theta(a|s) R(s,a) da ds = \mathbb{E}_{s \sim \rho^\pi, a \sim \pi_\theta}[R(s,a)] \quad (36)$$

where $\rho^\pi$ is the distribution of state transition. The objective gradient of PG $\nabla_\theta J(\theta) = \mathbb{E}_{\tau \sim \pi_\theta(\tau)}[\nabla_\theta \log \pi_\theta(\tau) R(\tau)]$ includes a vector $C = \nabla_\theta \log \pi_\theta(\tau)$ and a scalar $R = R(\tau)$. Vector $C$ is the trend of policy update, while scalar $R$ is range of this trend. Hence, the scalar $R$ acts as a critic that decides how policy is updated. Action value $Q^\pi(s,a)$ is defined as the expectation of discounted rewards by

$$Q^\pi(s,a) = \mathbb{E}[r_1^\gamma = \sum_{k=t}^\infty \gamma^{k-t} r(s_k, a_k) | S_1 = s, A_1 = a; \pi]. \quad (37)$$

$Q^\pi(s,a)$ is an alternative of scalar $R$, and it is better than $R$ as critic. Therefore, objective gradient of PG changes to

$$\nabla_\theta J(\theta) = \nabla_\theta \int_{s \sim S} \rho^\pi(s) \int_{a \sim A} \pi_\theta(a|s) Q^\pi(s,a) da ds = \mathbb{E}_{s \sim \rho^\pi, a \sim \pi_\theta}[\nabla_\theta \log \pi_\theta(a|s) Q^\pi(s,a)], \quad (38)$$

and policy is updated using objective gradient with action value $Q^\pi(s,a)$. Hence, algorithms are called ***stochastic policy gradient algorithm*** if action value $Q^\pi(s,a)$ is used as critic.

**DPG:** DPG are algorithms that train a deterministic policy $\mu_\theta(s)$ to select actions, instead of policy $\pi_\theta(a|s)$ in AC. A policy is *deterministic policy* $\mu_\theta(s)$ if it maps state to action $a \leftarrow \mu_\theta(s)$, while stochastic policy $\pi_\theta(a|s)$ maps state and action to a probability $P(a|s)$ [47]. The update of deterministic policy is defined as

$$\mu^{k+1}(s) = \arg\max_a Q^{\mu^k}(s,a). \quad (39)$$

If network $\theta$ is used as approximator of deterministic policy, update of network changes to

$$\theta^{k+1} = \theta^k + \alpha \mathbb{E}_{s \sim \rho^{\mu^k}}\left[\nabla_\theta Q^{\mu^k}(s, \mu_\theta(s))\right] = \theta^k + \alpha \mathbb{E}_{s \sim \rho^{\mu^k}}\left[\nabla_\theta \mu_\theta(s) \nabla_a Q^{\mu^k}(s,a)|_{a=\mu_\theta(s)}\right]. \quad (40)$$

There are small changes in state distribution $\rho^\mu$ of deterministic policy during the update of network $\theta$, but this change will not impact the update of network. Hence, network of deterministic policy is updated by

$$\theta \leftarrow \theta + \alpha \mathbb{E}_{s \sim \rho^\mu}\left[\nabla_\theta \mu_\theta(s) \nabla_a Q^\mu(s,a)|_{a=\mu_\theta(s)}\right] \quad (41)$$

because

$$\nabla_\theta J(\mu_\theta) = \nabla_\theta \int_S \rho^\mu(s) R(s, \mu_\theta(s)) ds = \nabla_\theta \mathbb{E}_{s \sim \rho^\mu}[R(s, \mu_\theta(s))] = \int_S \rho^\mu(s) \nabla_\theta \mu_\theta(s) \nabla_a Q^\mu(s,a)|_{a=\mu_\theta(s)} = \mathbb{E}_{s \sim \rho^\mu}\left[\nabla_\theta \mu_\theta(s) \nabla_a Q^\mu(s,a)|_{a=\mu_\theta(s)}\right]. \quad (42)$$

Once $Q^\mu(s,a)$ is obtained, $\theta$ can be updated after obtaining objective gradient.

How to find $Q^\mu(s,a)$? Note that discounted reward $Q^\mu(s,a)$ is a critic in stochastic policy gradient mentioned before. If network $w$ is used as approximator, $Q^\mu(s,a)$ is obtained by

$$Q^\mu(s,a) \approx Q^w(s,a) \quad (43)$$

stochastic policy gradient algorithm includes 2 networks, in which $w$ is the critic that approximates action value and $\theta$ is used as actor to select actions in test (actions are selected by behavior policy $\beta$ in training). Stochastic policy gradient in this case is called *off-policy deterministic actor-critic* (OPDAC) or *OPDAC-Q*. Objective gradient of OPDAC therefore changes from the Eq. 42 to

$$\nabla_\theta J_\beta(\mu_\theta) \approx \mathbb{E}_{s \sim \rho^\beta}\left[\nabla_\theta \mu_\theta(s) \nabla_a Q^w(s,a)|_{a=\mu_\theta(s)}\right] \quad (44)$$

where $\beta$ represents the behavior policy. 2 networks are updated by

$$\delta_t = r_t + \gamma Q^w(s_{t+1}, \mu_\theta(s_{t+1})) - Q^w(s_t, a_t) \quad (45)$$

$$w_t \leftarrow w_t + \alpha_w \delta_t \nabla_w Q^w(s_t, a_t) \quad (46)$$

$$\theta_t \leftarrow \theta_t + \alpha_\theta \nabla_\theta \mu_\theta(s) \nabla_a Q^w(s,a)|_{a=\mu_\theta(s)}. \quad (47)$$

However, no constrains is used on network $w$ in approximation process will lead to a large bias.

How to obtain a $Q^w(s,a)$ without bias? *Compatible function approximation* (CFA) can eliminate the bias by adding two requirements on $w$ (proof is given in [47]): (1) $\nabla_a Q^w(s,a)|_{a=\mu_\theta(s)} = \nabla_\theta \mu_\theta(s)^\mathsf{T} w$ ; (2) $MSE(\theta,w) = \mathbb{E}[\epsilon(s;\theta,w)^\mathsf{T}\epsilon(s;\theta,w)] \to 0$ where $\epsilon(s;\theta,w) = \nabla_a Q^w(s,a)|_{a=\mu_\theta(s)} - \nabla_a Q^\mu(s,a)|_{a=\mu_\theta(s)}$. In other words, $Q^w(s,a)$ should meet

$$Q^w(s,a) = (a - \mu_\theta(s))^\mathsf{T} \nabla_\theta \mu_\theta(s)^\mathsf{T} w + V^v(s) \quad (48)$$

where state value $V^v(s)$ may be any *differentiable baseline function* [47]. Here $v$ and $\emptyset$ are feature and parameter of state value ($V^v(s) = v^\mathsf{T}\emptyset(s)$). Parameter $\emptyset$ is also the feature of advantage function ($A^w(s,a) = \emptyset(s,a)^\mathsf{T} w$), and $\emptyset(s,a)$ is defined as $\emptyset(s,a) \stackrel{\text{def}}{=} \nabla_\theta \mu_\theta(s)(a - \mu_\theta(s))$. Hence, a low-bias $Q^w(s,a)$ is obtained using OPDAC-Q and CFA. This new algorithm with less bias is called *Compatible OPDAC-Q* (COPDAC-Q) [47], in which weights are updated as Eq. 49-51

$$v_t \leftarrow v_t + a_v \delta_t \emptyset(s_t) \quad (49)$$

$$w_t \leftarrow w_t + \alpha_w \delta_t \emptyset(s_t,a_t) = w_t + \alpha_w \delta_t \nabla_w A^w(s_t,a_t) \quad (50)$$

$$\theta_t \leftarrow \theta_t + \alpha_\theta \nabla_\theta \mu_\theta(s_t)(\nabla_\theta \mu_\theta(s_t)^\mathsf{T} w_t) \quad (51)$$

where $\delta_t$ is the same as the Eq. 45. Here $a_v$, $\alpha_w$ and $\alpha_\theta$ are learning rates. Note that *linear function approximation method* [47] is used to obtain advantage function $A^w(s,a)$ that is used to replace the value function $Q^w(s,a)$ because $A^w(s,a)$ is efficient than $Q^w(s,a)$ in weight update. Linear function approximation however may lead to divergence of $Q^w(s,a)$ in critic $\delta$. Critic $\delta$ can be replaced by the *gradient Q-learning critic* [52] to reduce divergence. Algorithm that combines COPDAC-Q and gradient Q-learning critic is called *COPDAC Gradient Q-learning* (COPDAC-GQ). Details of gradient Q-learning critic and COPDAC-GQ algorithm can be found in [47][52].

By analytical illustration above, 2 examples (COPDAC-Q and COPDAC-GQ) of DPG algorithm are obtained. In short, key points of DPG is to (1) find a no-biased $Q^w(s,a)$ as critic; (2) train a deterministic policy $\mu_\theta(s)$ to select actions. networks of DPG are updated as AC. Brief steps of DPG is shown in Fig. 23.

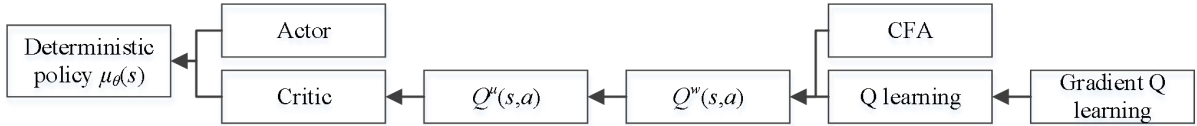

Figure 23. Brief steps of DPG algorithm.

**DDPG** [67] is the combination of replay buffer, deterministic policy $\mu(s)$ and actor-critic architecture. $\theta^Q$ is used as critic network to approximate action value $Q(s_i,a_i;\theta^Q)$. $\theta^\mu$ is used as policy network to approximate deterministic policy $\mu(s;\theta^\mu)$. TD target $y$ of DDPG is defined by

$$y_i = r_i + \gamma Q'(s_{i+1},\mu'(s_{i+1};\theta^{\mu'});\theta^{Q'}) \quad (52)$$

where $\theta^{Q'}$ and $\theta^{\mu'}$ are copies of $\theta^Q$ and $\theta^\mu$ as target networks that update with low frequency. The objective of critic network is defined by

$$J(\theta^Q) = y_i - Q(s_i,a_i;\theta^Q). \quad (53)$$

Critic network $\theta^Q$ is updated by minimizing the loss value (MSE loss)

$$Loss = \frac{1}{N}\sum_i J(\theta^Q)^2 \quad (54)$$

where $N$ is the number of tuples $<s,a,r,s'>$ sampled from replay buffer. Target function of policy network is defined by

$$J(\theta^\mu) = \frac{1}{N}\sum_i Q(s_i,a_i;\theta^\mu) \quad (55)$$

and objective gradient is obtained by

$$\nabla_{\theta^\mu} J(\theta^\mu) \cong \frac{1}{N}\sum_i \nabla_{\theta^\mu}\mu(s_i;\theta^\mu)\nabla_a Q(s_i,a;\theta^Q)|_{a=\mu(s_i)}. \tag{56}$$

Hence, policy network $\theta^\mu$ is updated according to gradient ascent method by

$$\theta^\mu \leftarrow \theta^\mu + \alpha\nabla_{\theta^\mu} J(\theta^\mu) \tag{57}$$

where $\alpha$ is a learning rate. New target networks

$$\theta^{Q'} \leftarrow \tau\theta^Q + (1-\tau)\theta^{Q'} \tag{58}$$

$$\theta^{\mu'} \leftarrow \tau\theta^\mu + (1-\tau)\theta^{\mu'} \tag{59}$$

where $\tau$ is a learning rate, are obtained by "soft" update method that improves the stability of network convergence. Detailed steps of DDPG are shown in **Algorithm 4** [67] and Fig. 24. Examples can be found in [27][46] in which DDPG is used in robotic arms.

---

**Algorithm 4:** DDPG

---

Randomly initialize critic network $Q(s,a;\theta^Q)$ and actor $\mu(s;\theta^\mu)$ with weight $\theta^Q$ and $\theta^\mu$
Initialize target network $Q'$ and $\mu'$ with weights $\theta^{Q'} \leftarrow \theta^Q$, $\theta^{\mu'} \leftarrow \theta^\mu$
Initialize replay buffer $\mathcal{R}$
**for** episode =1, $M$ **do**
  Initialize a random process $\mathcal{N}$ for action exploration
  Receive initial observation state $s_1$
  **for** t=1, $T$ **do**
    Selection action $a_t = \mu(s_t;\theta^\mu) + \mathcal{N}_t$ according to the current policy and exploration noise
    Execute action $a_t$ and observe reward $r_t$ and observe new state $s_{t+1}$
    Store transition $(s_t,a_t,r_t,s_{t+1})$ in $\mathcal{R}$
    Sample a random minibatch of $\mathcal{N}$ transitions $(s_i,a_i,r_i,s_{i+1})$ from $\mathcal{R}$
    Set $y_i = r_i + \gamma Q'(s_{i+1},\mu'(s_{i+1};\theta^{\mu'});\theta^{Q'})$
    Update critic by minimization the loss: $Loss = \frac{1}{N}\sum_i (y_i - Q(s_i,a_i;\theta^Q))^2$
    Update the actor policy using the sampled policy gradient:
$$\nabla_{\theta^\mu} J(\theta^\mu) \approx \frac{1}{N}\sum_i \nabla_{\theta^\mu}\mu(s_i;\theta^\mu)\nabla_a Q(s_i,a;\theta^Q)|_{a=\mu(s_i)}$$
    Update the target network:
$$\theta^{Q'} \leftarrow \tau\theta^Q + (1-\tau)\theta^{Q'}$$
$$\theta^{\mu'} \leftarrow \tau\theta^\mu + (1-\tau)\theta^{\mu'}$$
  **end for**
**end for**

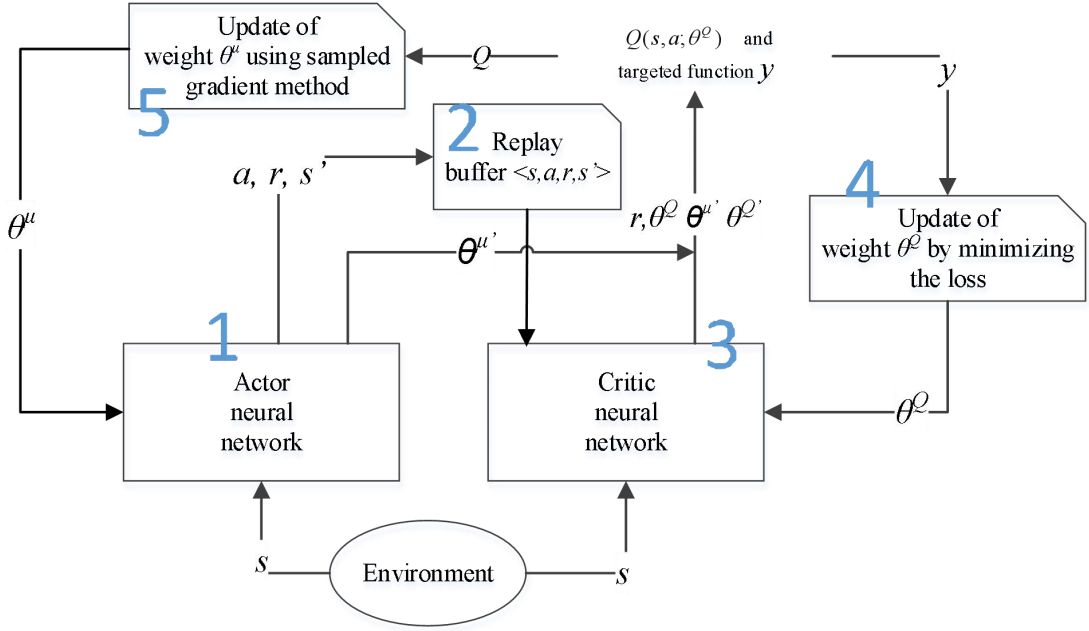

Figure 24. Steps of DDPG. DDPG combines the replay buffer, actor-critic architecture, and deterministic policy. First, action is selected by policy network and reward is obtained. State transits to next state. Second, experience tuple is saved in replay buffer. Third, experiences are sampled from replay buffer for training. Fourth, critic network is updated. Finally, policy network is updated.

## 5.5 TRPO and PPO

PPO [70][85] is the optimized version of TRPO [69]. Hence, here we first introduce TRPO, and then introduce PPO.

**TRPO:** Previous policy gradient algorithms update their policies by $\theta \leftarrow \theta + \nabla_\theta J(\theta)$. However, new policy is improved unstably. The goal of TRPO is to improve its policy monotonously, therefore stability of convergence is improved by finding a new policy with the objective that is defined by

$$J(\theta) = L_{\theta_{old}}(\theta), s.t. D_{KL}^{max}(\theta_{old},\theta) \leq \delta \qquad (60)$$

where $L_{\theta_{old}}(\theta)$ is the approximation of new policy's expectation, $D_{KL}^{max}(\theta_{old},\theta)$ the KL divergence and $\delta$ a *trust region constraint* of KL divergence. The objective gradient $\nabla_\theta J(\theta)$ is obtained by maximizing the objective $J(\theta)$.

$\eta(\theta)$ and $\eta(\theta_{old})$ denote expectations of new and old policies respectively. Their relationship is defined by $\eta(\theta) = \eta(\theta_{old}) + \mathbb{E}_{s_0,a_0,s_1,a_1\ldots}[\sum_{t=0}^{\infty} \gamma^t A_{\theta_{old}}(s_t,a_t)]$ where $\gamma$ is a discount factor, and $A_{\theta_{old}}(s_t,a_t)$ is the advantage value that is defined by $A_\theta(s,a) = Q_\theta(s,a) - V_\theta(s)$. Thus, $\eta(\theta) = \eta(\theta_{old}) + \sum_s \rho_\theta(s) \sum_a \theta(a|s) A_{\theta_{old}}(s,a)$ where $\rho_\theta(s)$ is the probability distribution of new policy, but $\rho_\theta(s)$ is unknown therefore it is impossible to obtain new policy $\eta(\theta)$. Approximation of new policy's expectation $L_{\theta_{old}}(\theta)$ is defined by

$$L_{\theta_{old}}(\theta) = \eta(\theta_{old}) + \sum_s \rho_{\theta_{old}}(s) \sum_a \theta(a|s) A_{\theta_{old}}(s,a) = \eta(\theta_{old}) + \sum_s \rho_{\theta_{old}}(s) \sum_a \frac{\theta(a|s)}{\theta_{old}(a|s)} \cdot \theta_{old}(a|s) \cdot A_{\theta_{old}}(s,a) = \eta(\theta_{old}) + \mathbb{E}[\frac{\theta(a|s)}{\theta_{old}(a|s)} A_{\theta_{old}}(s,a)] \quad (61)$$

where $\rho_{\theta_{old}}(s)$ is known. The relationship of $L_{\theta_{old}}(\theta)$ and $\eta(\theta)$ [78] is proved to be

$$\eta(\theta) \geq L_{\theta_{old}}(\theta) - C \cdot D_{KL}^{max}(\theta_{old},\theta) \qquad (62)$$

where penalty coefficient $C = \frac{2\epsilon\gamma}{(1-\gamma)^2}$, $\gamma \in [0,1]$ and $\epsilon$ the maximum advantage. Hence, it is possible to obtain $\eta(\theta)$ by $maximize_\theta [L_{\theta_{old}}(\theta) - C \cdot D_{KL}^{max}(\theta_{old},\theta)]$ or

$$maximize_\theta \ \mathbb{E}[\frac{\theta(a|s)}{\theta_{old}(a|s)}A_{\theta_{old}}(s,a) - C \cdot D_{KL}^{max}(\theta_{old},\theta)]. \tag{63}$$

However, penalty coefficient $C$ (constrain of KL divergence) will lead to small step size in policy update. A trust region constraint $\delta$ is used to constrain KL divergence by

$$maximize_\theta \ \mathbb{E}[\frac{\theta(a|s)}{\theta_{old}(a|s)}A_{\theta_{old}}(s,a)], \ s.t. \ D_{KL}^{max}(\theta_{old},\theta) \leq \delta \tag{64}$$

therefore step size in policy update is enlarged robustly. New improved policy is obtained in trust region by maximizing objective $L_{\theta_{old}}(\theta)$, s.t. $D_{KL}^{max}(\theta_{old},\theta) \leq \delta$. This objective can be simplified further [69] and new policy is obtained by

$$maximize_\theta \ [\nabla_\theta L_{\theta_{old}}(\theta)|_{\theta=\theta_{old}} \cdot (\theta - \theta_{old})]$$
$$s.t. \ \frac{1}{2}\|\theta - \theta_{old}\|^2 \leq \delta. \tag{65}$$

**PPO**: we know objective of TRPO is $maximize_\theta \ \mathbb{E}[\frac{\theta(a|s)}{\theta_{old}(a|s)}A_{\theta_{old}}(s,a)], \ s.t. \ D_{KL}^{max}(\theta_{old},\theta) \leq \delta$, in which a fixed trust region constraint $\delta$ is used to constrain KL divergence instead of penalty coefficient $C$. Fixed trust region constraint $\delta$ leads to a reasonable step size in policy update therefore stability in convergence is improved and convergence speed is acceptable. However, objective of TRPO is obtained in implementation by conjugate gradient method [70] that is computationally expensive.

PPO optimizes objective $maximize_\theta \ \mathbb{E}[\frac{\theta(a|s)}{\theta_{old}(a|s)}A_{\theta_{old}}(s,a) - C \cdot D_{KL}^{max}(\theta_{old},\theta)]$ from 2 aspects: (1) probability ratio $r(\theta) = \frac{\theta(a|s)}{\theta_{old}(a|s)}$ in objective is constrained in interval $[1-\epsilon, 1+\epsilon]$ by introducing *"surrogate" objective*

$$L^{CLIP}(\theta) = \mathbb{E}\{\min[r(\theta)A, clip(r(\theta), 1-\epsilon, 1+\epsilon)]\} \tag{66}$$

where $\epsilon$ is a hyperparameter, to penalize changes of policy that move $r(\theta)$ away from 1 [70]; (2) penalty coefficient $C$ is replaced by *adaptive penalty coefficient $\beta$* that increases or decreases according to the expectation of KL divergence in new update. To be exact,

$$if \ d < \frac{d_{targ}}{1.5}, \beta \leftarrow \frac{\beta}{2}; if \ d > d_{targ} \times 1.5, \beta \leftarrow \beta \times 2 \tag{67}$$

where $d = \mathbb{E}[\mathbb{D}_{KL}^{max}(\theta_{old},\theta)]$ and $d_{targ}$ denotes target value of KL divergence in each policy update, therefore *KL-penalized objective* is obtained by

$$L^{KLPEN}(\theta) = \mathbb{E}[\frac{\theta(a|s)}{\theta_{old}(a|s)}A_{\theta_{old}}(s,a) - \beta \cdot D_{KL}^{max}(\theta_{old},\theta)]. \tag{68}$$

In the implementation with neural network, loss function is required to combine the policy surrogate and value function error [70], and entropy are also used in objective to encourage exploration. Hence, *combined surrogate objective* is obtained by

$$L^{CLIP+VF+S}(\theta) = \mathbb{E}[L^{CLIP}(\theta) + c_1 L^{VF}(\theta) + c_2 S(\pi_\theta|s)] \tag{69}$$

where $c_1$, $c_2$, S and $L^{VF}(\theta)$ denote 2 coefficients, entropy bonus and square-error loss respectively. Objectives ($L^{CLIP+VF+S}(\theta)$ and $L^{KLPEN}(\theta)$) of PPO is optimized by SGD that cost less computing resource than conjugate gradient method. PPO is implemented with actor-critic architecture, therefore it converges faster than TRPO.

# VI. Analytical comparisons

To provide a clear understanding about advantages and disadvantages of different motion planning algorithms, we divide them into 4 groups: traditional algorithms, supervised learning algorithms, optimal value RL and policy gradient RL, and comparisons are made according to their principles mentioned in section II, III, IV and V. First, direct comparisons of algorithms in each group are made to provide a clear understanding about the input, output, and key features of these algorithms. Second, analytical comparisons of all motion planning algorithms are made to provide a comprehensive understanding about performance and application of algorithms, according to general criteria. Third, analytical comparisons about the convergence of RL-based motion planning algorithms are specially made, because RL-based algorithms are the research focus recently.

## 6.1 Direct comparisons of motion planning algorithms

**Traditional algorithms:** this group includes graph search algorithms, sampling-based algorithms, and interpolating curve algorithms. Table 1 lists their input, output and key features: (1) these algorithms use graph or map of workspace as input, and output trajectory directly; (2) graph search algorithms find shortest and collision-free trajectory by search methods (e.g. best-first search). For example, Dijkstra's algorithm is based on best-first search. However, search process is computationally expensive because search space is large, therefore heuristic function is used to reduce search space and the shortest path is found by estimating the overall cost (e.g. A*); (3) sampling-based algorithms randomly sample a collision-free trajectory in search space (e.g. PRM), and constraints (e.g. non-holonomic constraint) are needed for some algorithms (e.g. RRT) in sampling process; (4) interpolating curve algorithms plan their path by mathematical rules, and then planned path is smoothed by CAGD.

Table 1. Comparison of traditional planning algorithms.

| Classification | Example | input | Key features | output |
|---|---|---|---|---|
| Graph search algorithm | Dijkstra's [1] | Graph or map | 1. Best-first search (large search space) 2. Heuristic function in cost estimation | trajectory |
| | A* [1,2] | | | |
| Sampling based algorithm | PRM [1] | | 1. Random search (suboptimal path) 2. Non-holonomic constraint | |
| | RRT [1,2] | | | |
| Interpolating Curve algorithm | Line and circle | | Mathematical rules; Path smoothing using CAGD | |
| | Clothoid curves | | | |
| | Polynomial curves | | | |
| | Bezier curves | | | |
| | Spline curves | | | |

**Supervised learning algorithms:** this group includes MSVM, LSTM, MCTS and CNN. These algorithms are listed in Table 2: (1) MSVM, LSTM and MCTS use well-prepared vector as input, while CNN can directly use image as input; (2) LSTM and MCTS can output time-sequential actions, because of their structures (e.g. tree) that can store and learn time-sequential features. MSVM and CNN cannot output time-sequential actions because they output one-step prediction by performing trained classifier; (3) MSVM plans the motion of robots by training a maximum margin classifier. LSTM stores and processes inputs in its cell that is a stack structure, and then actions are outputted by performing trained LSTM model. MCTS is the combination of Monte-carlo method and search tree. Environmental states and values are stored and updated in its node of tree, therefore actions are outputted by performing trained MCTS model. CNN converts high-dimensional images to low-dimensional features by convolutional layers. These low-dimensional features are used to train a CNN model, therefore actions are outputted by performing trained CNN model.

Table 2. Comparison of supervised learning algorithms.

| Algorithm | Input | Key features | output |
|---|---|---|---|
| MSVM | Vector | Maximum margin classifier | None-sequential actions |
| LSTM | Vector | Cell (stack structure) | Time-sequential actions |
| MCTS | Vector | Monte-carlo method; Tree structure | Time-sequential actions |
| CNN | Image | Convolutional layers; Weight matrix | None-sequential actions |

**Optimal value RL:** this group here includes Q learning, DQN, double DQN, and dueling DQN. Features of algorithms here include replay buffer, objectives of algorithm, and Weight update method. Comparisons of these algorithms are listed in Table 3: (1) Q learning normally uses well-prepared vector as input, while DQN, double DQN and dueling DQN use images as input because these algorithms use convolutional layer to process high-dimensional images; (2) outputs of these algorithms are time-sequential actions by performing trained model; (3) DQN, double DQN and dueling DQN use replay buffer to reuse experience, while Q learning collects experiences and learns from then in an online way; (4) DQN, double DQN and dueling DQN use MSE $e^2$ as their objectives. Their differences are: first, DQN obtains action value by neural network $Q(s,a;\theta)$, while Q learning obtains action value by querying Q-table; second, double DQN use another neural network $\theta^-$ to evaluate selected action to obtain a better action value by $Q(s',\arg max_a Q(s',a';\theta');\theta^-)$; third, dueling DQN obtains action value by dividing action value to advantage value and state value. Constraint $\mathbb{E}_{a\sim\pi(s)}[A(s,a)] = 0$ is used on advantage value, therefore a better action value is obtained by $Q(s,a;\theta,\alpha,\beta) = V(s;\theta,\beta) + \{A(s,a;\theta,\alpha) - \frac{1}{|\mathcal{A}|}\sum_{a_t^- \in \mathcal{A}} A(s_t, a_t^-;\theta,\alpha)\}$. Networks that approximate action value in these algorithms are updated by minimizing MSE with gradient descent approach.

Table 3. Comparison of optimal-value RL.

| Algorithm | Input | output | Replay buffer | Objective | Weight update method |
|---|---|---|---|---|---|
| Q learning | Vector | Time-sequential actions | No | $e^2$ where $e = r + \gamma max_a Q(s',a') - Q(s,a)$ | gradient descent |
| DQN | Image | Time-sequential actions | Yes | $e^2$ where $e = r + \gamma max_a Q(s',a';\theta') - Q(s,a;\theta)$ | gradient descent |
| Double DQN | Image | Time-sequential actions | Yes | $e^2$ where $e = r + \gamma Q(s', \arg max_a Q(s',a';\theta'); \theta^-) - Q(s,a;\theta)$ | gradient descent |
| Dueling DQN | Image | Time-sequential actions | Yes | $e^2$ where $e = r + \gamma max_a Q(s',a';\theta) - Q(s,a;\theta)$ and $Q(s,a;\theta,\alpha,\beta) = V(s;\theta,\beta) + \left\{A(s,a;\theta,\alpha) - \frac{1}{|\mathcal{A}|}\sum_{a_t^- \in \mathcal{A}} A(s_t, a_t^-;\theta,\alpha)\right\}$ | gradient descent |

**Policy gradient RL:** this group here includes PG, AC, A3C, A2C, DPG, DDPG, TRPO, and PPO. Features of these algorithms include actor-critic architecture, multi-thread method, replay buffer, objective of algorithm, and weight update method. Comparisons of these algorithms are listed in Table 4: (1) input of policy gradient RL can be image or vector, and image is used as inputs under the condition that convolutional layer is used as preprocessing component to convert high-dimensional image to low-dimensional feature; (2) outputs of policy gradient RL are time-sequential actions by performing trained policy $\pi(s): s \to a$; (3) actor-critic architecture is not used in PG, while other policy gradient RL are implemented with actor-critic architecture; (4) A3C and A2C use multi-thread method to collect data and update their network, while other policy gradient RL are based on single thread in data collection and network update; (5) DPG and DDPG use replay buffer to reuse data in an offline way, while other policy gradient RL learn online; (6) the objective of PG is defined as the expectation of accumulative rewards in trajectory by $\mathbb{E}_{\tau \sim \pi_\theta(\tau)}[R(\tau)]$. Critic objectives of AC, A3C, A2C, DPG and DDPG are defined as MSE $e^2$, and their critic networks are updated by minimizing the MSE. However, their actor objectives are different because: first, actor objective of AC is defined as $\pi_\theta(a_t|s_t) \cdot e$; second, policy entropy is added on $\pi_\theta(a_t|s_t) \cdot e$ to encourage exploration, therefore actor objectives of A3C and A2C are defined by $\pi_\theta(a_t|s_t) \cdot e + \beta H(\pi(s_t;\theta))$; third, DPG and DDPG use action value as their actor objectives by $Q(s_i,a_i;\theta^\mu)$ that are approximated by neural network, and their policy networks (actor networks) are updated by obtaining objective gradient $\nabla_{\theta^\mu}\mu(s_i;\theta^\mu)\nabla_a Q(s_i,a;\theta^Q)|_{a=\mu(s_i)}$. Critic objectives of TRPO and PPO are defined as the advantage value by $A_t = \delta_t + (\gamma\lambda)\delta_{t+1} + \ldots + \gamma\lambda^{T-t}\delta(s_T)$ where $\delta_t = r_t + \gamma V(s_{t+1};w) - V(s_t;w)$, and their critic networks $w$ are updated by minimizing $\delta_t^2$. Actor objectives of TRPO and PPO are different: objective of TRPO is defined as $\mathbb{E}[\frac{\theta(a|s)}{\theta_{old}(a|s)}A_{\theta_{old}}(s,a)]$, s.t. $D_{KL}^{max}(\theta_{old},\theta) \leq \delta$ in which a fixed trust region constraint $\delta$ is used to ensure the monotonous update of policy network $\theta$, while PPO uses "surrogate" $[1-\epsilon, 1+\epsilon]$ and adaptive penalty $\beta$ to ensure a better monotonous update of policy network, therefore PPO has 2 objectives that are defined as $L^{KLPEN}(\theta) = \mathbb{E}[\frac{\theta(a|s)}{\theta_{old}(a|s)}A_{\theta_{old}}(s,a) - \beta \cdot D_{KL}^{max}(\theta_{old},\theta)]$ (KL-penalized objective) and $L^{CLIP+VF+S}(\theta) = \mathbb{E}[L^{CLIP}(\theta) + c_1 L^{VF}(\theta) + c_2 S(\pi_\theta|s)]$ (surrogate objective) where $L^{CLIP}(\theta) = \mathbb{E}\{\min[r(\theta)A, clip(r(\theta), 1-\epsilon, 1+\epsilon)]\}$; (7) policy network of policy gradient RL are all updated by gradient ascent approach $\theta \leftarrow \theta + \alpha \nabla_\theta J(\theta)$. Policy networks of A3C and A2C are updated in asynchronous and synchronous ways respectively. Networks of DDPG is updated in a "soft" way by $\theta^{Q'} \leftarrow \tau\theta^Q + (1-\tau)\theta^{Q'}$ and $\theta^{\mu'} \leftarrow \tau\theta^\mu + (1-\tau)\theta^{\mu'}$.

Table 4. Comparison of policy gradient RL.

| Algorithm | Input | output | Actor-critic architecture | Multi-thread method | Replay Buffer | Objective | Weight update method |
|---|---|---|---|---|---|---|---|
| PG | Image/ vector | Time-sequential actions | ---* | --- | --- | 1 | Gradient ascent |
| AC | Image/ vector | Time-sequential actions | Yes | --- | --- | Critic: 2 Actor: 3 | Gradient ascent |
| A3C | Image/ vector | Time-sequential actions | Yes | Yes | --- | Critic: 4 Actor: 5 | Gradient ascent Asynchronous update |
| A2C | Image/ vector | Time-sequential actions | Yes | Yes | --- | Critic: 6 Actor: 7 | Gradient ascent Synchronous update |
| DPG | Image/ vector | Time-sequential actions | Yes | --- | Yes | Critic: 8 Actor: 9 | Gradient ascent |
| DDGP | Image/ vector | Time-sequential actions | Yes | --- | Yes | Critic: 10 Actor: 11 | Gradient ascent Soft update |
| TRPO | Image/ vector | Time-sequential actions | Yes | --- | --- | Critic: 12 Actor: 13 | Gradient ascent |
| PPO | Image/ vector | Time-sequential actions | Yes | --- | --- | Critic: 14 Actor: 15 | Gradient ascent |

*Here the mark "---" denotes "No".

1. $\mathbb{E}_{\tau \sim \pi_\theta(\tau)}[R(\tau)]$
2. $e^2, e = r_t + V(s_{t+1};w) - V(s_t;w)$
3. $\pi_\theta(a_t|s_t) \cdot e$
4. $e^2, e = \sum_{i=t}^{T} \gamma^{i-t} r_i + V(s_{t+n};w) - V(s_t;w)$
5. $\pi_\theta(a_t|s_t) \cdot e + \beta H(\pi(s_t;\theta))$
6. $e^2, e = \sum_{i=t}^{T} \gamma^{i-t} r_i + V(s_{t+n};w) - V(s_t;w)$
7. $\pi_\theta(a_t|s_t) \cdot e + \beta H(\pi(s_t;\theta))$

8. $e^2$, $e = r_t + \gamma Q^w(s_{t+1}, \mu_\theta(s_{t+1})) - Q^w(s_t, a_t)$

9. $Q^w(s_t, a_t)$, objective gradient: $\nabla_\theta \mu_\theta(s) \nabla_a Q^w(s,a)|_{a=\mu_\theta(s)}$

10. $e^2$, $e = r_i + \gamma Q'\left(s_{i+1}, \mu'\left(s_{i+1}; \theta^{\mu'}\right); \theta^{Q'}\right) - Q(s_i, a_i; \theta^Q)$

11. $Q(s_i, a_i; \theta^\mu)$, objective gradient: $\nabla_{\theta^\mu} \mu(s_i; \theta^\mu) \nabla_a Q(s_i, a; \theta^Q)|_{a=\mu(s_i)}$

12. $A_t = \delta_t + (\gamma\lambda)\delta_{t+1} + \ldots + \gamma\lambda^{T-t}\delta(s_T)$, $\delta_t = r_t + \gamma V(s_{t+1}; w) - V(s_t; w)$

13. $\mathbb{E}[\frac{\theta(a|s)}{\theta_{old}(a|s)} A_{\theta_{old}}(s,a)]$, s.t. $D_{KL}^{max}(\theta_{old}, \theta) \leq \delta$

14. $A_t = \delta_t + (\gamma\lambda)\delta_{t+1} + \ldots + \gamma\lambda^{T-t}\delta(s_T)$, $\delta_t = r_t + \gamma V(s_{t+1}; w) - V(s_t; w)$

15. (1) $L^{KLPEN}(\theta) = \mathbb{E}[\frac{\theta(a|s)}{\theta_{old}(a|s)} A_{\theta_{old}}(s,a) - \beta \cdot D_{KL}^{max}(\theta_{old}, \theta)]$

    (2) $L^{CLIP+VF+S}(\theta) = \mathbb{E}[L^{CLIP}(\theta) + c_1 L^{VF}(\theta) + c_2 S(\pi_\theta|s)]$

    where $L^{CLIP}(\theta) = \mathbb{E}\{\min[r(\theta)A, clip(r(\theta), 1-\epsilon, 1+\epsilon)]\}$

## 6.2 Analytical comparisons of motion planning algorithms

Here analytical comparisons of motion planning algorithms are made according to *general criteria* we summarized. These criteria include (1) local or global planning; (2) path length; (3) optimal velocity; (4) reaction speed; (5) safe distance; (6) time-sequential path. Speed and stability of network convergence for optimal value RL and policy gradient RL are then compared analytically because convergence speed and stability of RL in motion planning are recent research focus.

### A. comparisons according to general criteria

**Local or global planning:** this criteria denotes the area where the algorithm is used in most case. Table 5 lists planning algorithms and which criteria they fit: (1) graph search algorithms plan their path globally by search methods (e.g. depth-first search, best-first search) to obtain a collision-free trajectory on graph or map; (2) sampling-based algorithms samples local or global workspace by sampling methods (e.g. random tree) to find a collision-free trajectory; (3) interpolating curve algorithms draw fixed and short trajectory by mathematical rules to avoid local obstacles; (4) MSVM and CNN make one-step prediction by trained classifiers to decides their local motion; (5) LSTM, MCTS, optimal value RL and policy gradient RL can make time-sequential motion planning from the start to destination by performing their trained models. These models include stack structure model of LSTM, tree model of MCTS and matrix weight model of RL. These algorithms fit global motion planning tasks theoretically if size of workspace is not large, because it is hard to train a converged model in large workspace. In most case, models of these algorithms are trained in local workspace to make time-sequential prediction by performing their trained model or policy $\pi(s): s \to a$.

**Path length:** this criteria denotes the length of planned path that is described as "optimal path", "suboptimal path", and "fixed path". Path length of algorithms are listed in Table 5: (1) graph search algorithms can find a shortest path by performing search methods (e.g. best-first search) in graph or map; (2) sampling-based algorithms plan a suboptimal path. Their sampling method (e.g. random tree) leads to insufficient sampling that only covers a part of cases and suboptimal path is obtained; (3) interpolating curve algorithms plan their path according to mathematical rules that lead to a fixed length of path; (4) supervised learning algorithms (MSVM, LSTM, MCTS and CNN) plan their path by performing models that are trained with human-labeled dataset, therefore suboptimal path is obtained; (5) RL algorithms (optimal value RL and policy gradient RL) can generate optimal path under the condition that reasonable penalty is used to punish moved steps in the training, therefore optimal path is obtained by performing trained RL policy.

**Optimal velocity:** this criteria denotes the ability to tune the velocity when algorithms plan their path, therefore robot can reach the destination with minimum time along planned path. This criteria is described as "optimal velocity" and "suboptimal velocity". Table 5 lists performance of algorithms: (1) performance of graph search algorithms, sampling-based algorithms and interpolating algorithms in velocity tuning cannot be evaluated, because these algorithms are only designed for path planning to find a collision-free trajectory; (2) supervised learning algorithms (MSVM, LSTM, MCTS and CNN) can output actions that are in the format $\boldsymbol{v} = [v_x, v_y]$ where $v_x$ and $v_y$ are velocity in x and y axis, if algorithms are trained with these vector labels. However, these velocity-related labels are all hard-coded artificially. Time to reach destination heavily relies on artificial factor, therefore supervised learning algorithms cannot realize optimal velocity; (3) optimal value RL and policy gradient RL can realize optimal velocity by attaching penalty to consumed time in the training. These algorithms can automatically learn how to choose the best velocity in the training to cost time as less as possible, therefore robots can realize optimal velocity by performing trained policy. Note that in this case, actions in optimal value RL and policy gradient RL must be in format of $[v_x, v_y]$ and action space that contains many action choices must be defined.

**Reaction speed:** this criteria denotes the speed of reaction to dynamic obstacles. Reaction speed is described as 3 levels that includes "slow", "medium" and "fast". Table 5 lists reaction speed of algorithms: (1) graph search algorithms and sampling-based algorithms rely on planned trajectory in the graph or map to avoid obstacles. However, the graph or map is updated in a slow frequency normally, therefore reaction speed of these algorithms is slow; (2) interpolating curve algorithms plan their path according to mathematical rules that cost limited and predictable time in computation, therefore reaction speed of these algorithms is medium; (3) supervised learning algorithms, optimal value RL and policy gradient RL react to obstacles by performing trained model or policy $\pi(s): s \rightarrow a$ that maps state of environment to a probability distribution $P(a|s)$. This process is fast and time cost can be ignored, therefore reaction speed of these algorithms is fast.

**Safe distance:** this criteria denotes the ability to keep a safe distance to obstacles. Safe distance is described as 3 level that includes "fixed distance", "suboptimal distance" and "optimal distance". Table 5 lists the performance of algorithms: (1) graph search algorithms and sampling-based algorithms keep a fixed distance to static obstacles by hard-coded setting in robotic application. However, high collision rate is inevitable in dynamic environment because of slow update frequency of graph or map; (2) interpolating algorithms keep a fixed distance to static and dynamic obstacles according to mathematical rules; (3) supervised learning algorithms keep a suboptimal distance to static and dynamic obstacles. Suboptimal distance is obtained by performing a model that is trained with human-labeled dataset; (4) optimal value RL and policy gradient RL keep an optimal distance to static and dynamic obstacles by performing a trained policy $\pi(s): s \rightarrow a$. This policy is trained under the condition that penalty is used to punish close distance between robot and obstacles in the training, therefore algorithms will automatically learn how to keep an optimal distance to obstacles when robot moves towards destination.

**Time-sequential path:** this criteria denote whether an algorithm fits time-sequential task or not. Table 5 lists algorithms that fit time-sequential planning: (1) graph search algorithms, sampling-based algorithms and interpolating curve algorithms plan their path according to graph, map or mathematical rules, regardless of environment state in each time step. Hence, these algorithms cannot fit time-sequential task; (2) MSVM and CNN output actions by one-step prediction that has no relation with environment state in each time step; (3) LSTM and MCTS store environment state in each time step in their cells and nodes respectively, and their models are updated by learning from these time-related experience. Time-sequential actions are outputted by performing trained models, therefore these algorithms fit time-sequential task; (4) optimal value RL and policy gradient RL train their policy network by learning from environmental state in each time step. Time-sequential actions are outputted by performing trained policy, therefore these algorithms fit time-sequential task.

Table 5. Analytical comparisons according to general criteria.

| Algorithm | Local/global | path length | Optimal velocity | Reaction speed | Safe distance | Time-sequential path |
|---|---|---|---|---|---|---|
| Graph search alg. | Global | Optimal path | --* | Slow | Fixed distance; high Collison rate | No |
| Sampling-based alg. | Local; Global | Suboptimal path | -- | Slow | Fixed distance; high Collison rate | No |
| Interpolating curve alg. | Local | Fixed path | -- | Medium | Fixed distance | No |
| MSVM | Local | Suboptimal path | Suboptimal velocity | Fast | Suboptimal distance | No |
| LSTM | Local; Global | Suboptimal path | Suboptimal velocity | Fast | Suboptimal distance | Yes |
| MCTS | Local; Global | Suboptimal path | Suboptimal velocity | Fast | Suboptimal distance | Yes |
| CNN | Local | Suboptimal path | Suboptimal velocity | Fast | Suboptimal distance | No |
| Q learning | Local; Global | Optimal path | Optimal velocity | Fast | Optimal distance | Yes |
| DQN | Local; Global | Optimal path | Optimal velocity | Fast | Optimal distance | Yes |
| Double DQN | Local; Global | Optimal path | Optimal velocity | Fast | Optimal distance | Yes |
| Dueling DQN | Local; Global | Optimal path | Optimal velocity | Fast | Optimal distance | Yes |
| PG | Local; Global | Optimal path | Optimal velocity | Fast | Optimal distance | Yes |
| AC | Local; Global | Optimal path | Optimal velocity | Fast | Optimal distance | Yes |
| A3C | Local; Global | Optimal path | Optimal velocity | Fast | Optimal distance | Yes |
| A2C | Local; Global | Optimal path | Optimal velocity | Fast | Optimal distance | Yes |
| DPG | Local; Global | Optimal path | Optimal velocity | Fast | Optimal distance | Yes |
| DDGP | Local; Global | Optimal path | Optimal velocity | Fast | Optimal distance | Yes |
| TRPO | Local; Global | Optimal path | Optimal velocity | Fast | Optimal distance | Yes |
| PPO | Local; Global | Optimal path | Optimal velocity | Fast | Optimal distance | Yes |

*The mark "--" denotes the performance that cannot be evaluated.

**B. comparisons of convergence speed and stability**

**Convergence speed:** here we use "poor", "reasonable", "good", and "excellent" to describe the performance of convergence speed. Table 6 lists the performance of optimal value RL and policy gradient RL: (1) Q learning only fits simple motion planning with small-size Q table. It is hard to converge for Q learning with large-size Q table in complex environment. Over-estimation of Q value also leads to poor performance of Q learning if neural network is used to approximate Q value; (2) DQN suffers over-estimation of action value, but DQN learns from experience in replay buffer that make network reuse high-quality experience efficiently. Hence, convergence speed of DQN is reasonable; (3) double DQN uses another network $\theta^-$ to evaluate actions that are selected by $\theta'$. New action value with less over-estimation is obtained by $Q(s', \arg max_a Q(s',a';\theta'); \theta^-)$, therefore convergence speed is improved; (4) dueling DQN finds better action value by: first, dividing action value to state value and advantage value $Q(s,a) = V(s,a) + A(s,a)$; second, constraining advantage value $A(s,a)$ by $\mathbb{E}_{a \sim \pi(s)}[A(s,a)] = 0$, therefore new action value is obtained by $Q(s,a;\theta,\alpha,\beta) = V(s;\theta,\beta) + \{A(s,a;\theta,\alpha) - $ $\frac{1}{|\mathcal{A}|}\sum_{a_t^- \in \mathcal{A}} A(s_t, a_t^-; \theta, \alpha)\}$. Hence, performance of dueling DQN in convergence speed is good; (5) PG updates its policy according to trajectory rewards by $\mathbb{E}_{\tau \sim \pi_\theta(\tau)}[R(\tau)]$, therefore poor performance in convergence speed is inevitable; (5) AC uses critic network to evaluate actions selected by actor network, therefore speeding up the convergence; (6) A3C and A2C use multi-thread method to improve convergence speed directly, and policy entropy is also used to encourage exploration. These methods indirectly enhance the convergence speed; (7) performance of DPG and DDPG in convergence speed is good because: first, their critics are unbiased critic networks obtained by CFA and gradient Q learning; second, their policies are deterministic policy $\mu_\theta(s)$ that is faster than stochastic policy $\pi_\theta(a_t|s_t)$ in convergence speed; third, they update their networks offline with replay buffer; fourth, noise is used in DDPG to encourage exploration; (8) TRPO makes a great improvement in convergence speed by adding trust region constraint to policies by $D_{KL}^{max}(\theta_{old}, \theta) \leq \delta$, therefore its networks are updated monotonously by maximizing its objective $\mathbb{E}[\frac{\theta(a|s)}{\theta_{old}(a|s)} A_{\theta_{old}}(s,a)]$, s.t. $D_{KL}^{max}(\theta_{old}, \theta) \leq \delta$; (9) PPO moves further in improving convergence speed by introducing "surrogate" objective $L^{CLIP+VF+S}(\theta) = \mathbb{E}[L^{CLIP}(\theta) + c_1 L^{VF}(\theta) + c_2 S(\pi_\theta|s)]$ and KL-penalized objective $L^{KLPEN}(\theta) = \mathbb{E}[\frac{\theta(a|s)}{\theta_{old}(a|s)} A_{\theta_{old}}(s,a) - \beta \cdot D_{KL}^{max}(\theta_{old}, \theta)]$.

Table 6. comparison of speed in convergence for RL.

| Algorithm | Speed of convergence | Reasons |
|---|---|---|
| Q learning | Poor | Over-estimation of action value |
| DQN | Reasonable | Replay buffer |
| Double DQN | Good | Replay buffer; Another network for evaluation |
| Dueling DQN | Good | Replay buffer; Division of action value $Q(s,a) = V(s,a) + A(s,a)$ |
| PG | Poor | High variance by $\mathbb{E}_{\tau \sim \pi_\theta(\tau)}[R(\tau)]$ |
| AC | Reasonable | Actor-critic architecture |
| A3C | Good | Actor-critic architecture; Multi-thread method; Policy entropy |
| A2C | Good | Actor-critic architecture; Multi-thread method; Policy entropy |
| DPG | Good | Replay buffer; Actor-critic architecture; Deterministic policy; Unbiased critic network |
| DDGP | Good | Replay buffer; Actor-critic architecture; Deterministic policy; Unbiased critic network; Exploration noise |
| TRPO | Good | Actor-critic architecture; Fixed trust region constraint; |
| PPO | Excellent | Actor-critic architecture; "surrogate" objective; KL-penalized objective |

**Convergence stability:** Table 7 lists convergence stability of optimal value RL and policy gradient RL: (1) Q learning update its action value every step, therefore bias is introduced. Over-estimation of action value leads to suboptimal update direction of Q value of network is used as approximator. Hence, convergence stability of Q learning is poor; (2) DQN improves the convergence stability by replay buffer in which a batch of experiences are sampled and its network is update according to batch loss; (3) double DQN and dueling DQN find a better action value than DQN by evaluation network and advantage network respectively, therefore networks of these algorithms are updated towards a better direction; (4) PG updates its network according to trajectory reward. This reduces bias caused by one-step rewards, but introduce high variance. Hence, network of PG is updated with stability but it is still hard to converge; (5) performances of actor and critic network of AC is poor in early-stage training. This leads to a fluctuated update of networks in the beginning, although network is updated by gradient ascent approach $\theta \leftarrow \theta + \nabla_\theta J(\theta)$; (6) A3C and A2C update their networks by multi-step rewards $\sum_{i=t}^{T}\gamma^{i-t} r_i$ that reduces the bias and improves convergence stability, although it will introduce some variance. Gradient ascent approach also helps in convergence stability, therefore performance in convergence stability is reasonable; (6) unbiased critic, gradient ascent approach and replay buffer contribute to good performance in convergence stability for DPG and DDPG. Additionally, networks of DDPG are updated in a "soft" way by  $\theta^{Q'} \leftarrow \tau\theta^Q + (1-\tau)\theta^{Q'}$ and $\theta^{\mu'} \leftarrow \tau\theta^\mu + (1-\tau)\theta^{\mu'}$ that also contributes to convergence stability; (7) networks of TRPO and PPO is updated monotonously. TRPO achieve this goal by trust region constraints $D_{KL}^{max}(\theta_{old},\theta) \leq \delta$, while PPO uses "surrogate" objective $L^{CLIP+VF+S}(\theta) = \mathbb{E}[L^{CLIP}(\theta) + c_1 L^{VF}(\theta) + c_2 S(\pi_\theta|s)]$ and KL-penalized objective $L^{KLPEN}(\theta) = \mathbb{E}[\frac{\theta(a|s)}{\theta_{old}(a|s)} A_{\theta_{old}}(s,a) - \beta \cdot D_{KL}^{max}(\theta_{old},\theta)]$ . Hence, performance of their networks in convergence stability is good.

Table 7. comparison of stability in convergence for RL.

| Algorithms | Stability of convergence | Reasons |
| --- | --- | --- |
| Q learning | Poor | Update in each step; Over-estimation of action value |
| DQN | Reasonable | Replay buffer |
| Double DQN | Reasonable | Replay buffer; Evaluation network |
| Dueling DQN | Reasonable | Replay buffer; Advantage network |
| PG | Good | Update by trajectory rewards; Gradient ascent update |
| AC | Poor (in the beginning) | Update in each step; Gradient ascent update |
| A3C | Reasonable | Update by multi-step rewards; Gradient ascent update |
| A2C | Reasonable | Update by multi-step rewards; Gradient ascent update |
| DPG | Good | Unbiased critic; Gradient ascent update; Replay buffer |
| DDGP | Good | Unbiased critic; Gradient ascent update; Replay buffer; Soft update |
| TRPO | Good | Monotonous update; Gradient ascent update |
| PPO | Good | Monotonous update; Gradient ascent update |

# VII. Future directions of robotic motion planning

Here we first introduce a common but complex real-world motion planning task: **how to realize long-distance motion planning with safety and efficiency** (e.g. long-distance luggage delivery by robots)? Then research questions and directions are obtained by analyzing this task according to processing steps that include data collection, data preprocessing, motion planning and decision making (Fig. 25).

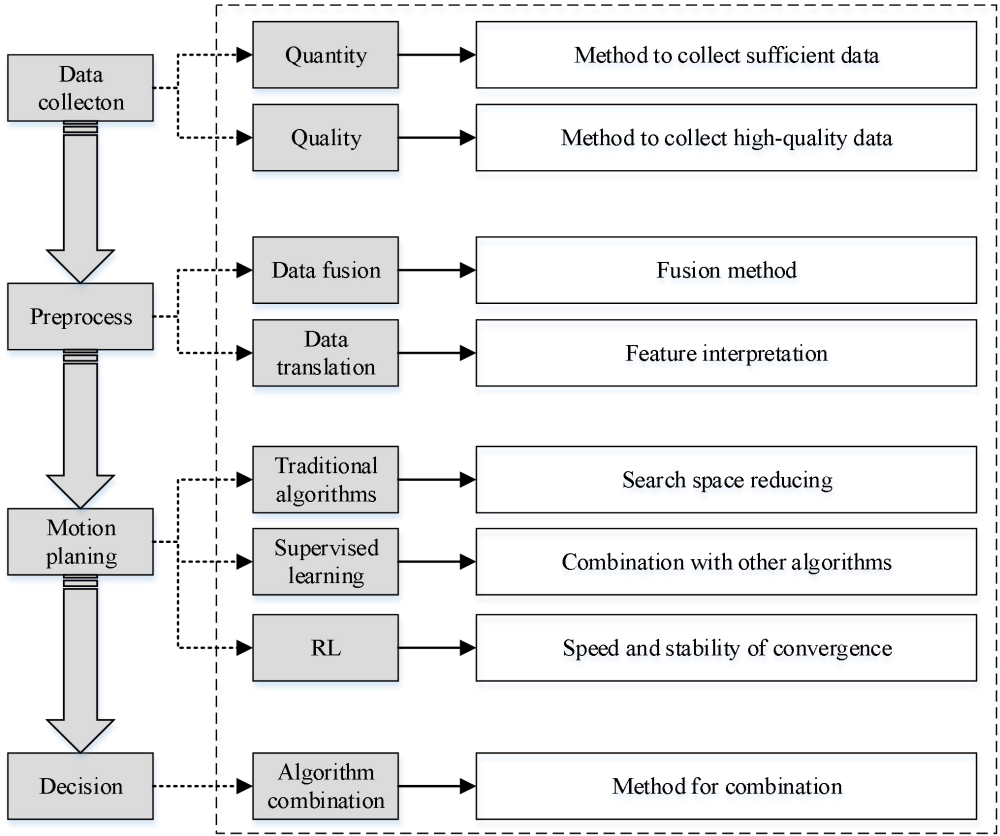

Figure 25. processing steps for motion planning task.

**Data collection:** to realize mentioned task, we may first consider: (1) how to collect enough data? (2) how to collect high-quality data? To collect enough data in a short time, we can consider collecting data by multi-thread method or cloud technology. Existing techniques seem enough to solve this question well. To collect high-quality data, existing works use prioritized replay buffer [80] to reuse high-quality data to train network. Imitation learning [79-80] is also used to collect high-quality data for network initialization, therefore network can converge faster (e.g. deep V learning [81-82]). Existing methods in data collection work well, therefore it is hard to make further optimization.

**Data preprocess: data fusion** and **data translation** should be considered after data is obtained. Multi-sensor data fusion algorithms [84] fuse data that is collected from same or different type of sensors. Data fusion is realized from pixel, feature, and decision levels, therefore partial understanding of environment is avoided. Another way to avoid partial understanding of environment is data translation that interpretate data to new format, therefore algorithms can have a better understanding about the relationship of robots and other obstacles (e.g. attention weight [82] and relation graph [83]). However, algorithms in data fusion and translation cannot fit all cases, therefore further works is needed according to the environment of application.

**Motion planning:** in this step, selection and optimization of motion planning algorithms should be considered: (1) if traditional motion planning algorithms (e.g. A*, RRT) are selected for task mentioned before, topological or global trajectory from the start to destination will be obtained, but this process is computationally expensive because of large search space. To solve this problem, the **combination of traditional algorithms and other ML algorithms** (e.g. CNN, DQN) may be a good choice. For example, RRT can be combined with DQN (Fig. 26) by using action value to predict directions of tree expansion, instead of heuristic or random search. (2) it seems impossible to use supervised learning to realize task mentioned above safely and quickly. Topological path is impossible to obtain by supervised learning that outputs one-step prediction. (3) topological path cannot be obtained by optimal value RL or policy gradient RL, but their performance in safety and efficiency is good locally by performing trained RL policy that leads to quick reaction, safe distance with obstacles, and shortest path or time. However, it is **time-consuming to train a RL policy because of deficiencies in network convergence**. Existing works made some optimizations to improve convergence (e.g. DDPG, PPO) in games to shorten training time of RL, but there is still a long way to go in real-world application. Recent trend to improve convergence is to create hybrid architecture that is the fusion of high-performance components (e.g. replay buffer, actor-critic architecture, policy entropy, multi-thread method).

**Decision:** traditional algorithms (e.g. A*) feature topological trajectory planning, while optimal value RL and policy gradient RL feature safe and quick motion planning locally. It is a good match to realize task mentioned above, by combining traditional algorithm with RL. Hence, overall robotic path is expected to approximate shortest path, and safety and efficiency can be ensured simultaneously. However, it is an engineering work, instead of research work.

To conclude, Fig. 25 lists possible research directions, but attentions to improve the performance of robotic motion planning are expected to be: (1) data fusion and translation of inputted features; (2) the optimization in traditional planning algorithms to reduce search space by combining traditional algorithms with supervised learning or RL; (3) the optimization in network convergence for RL.

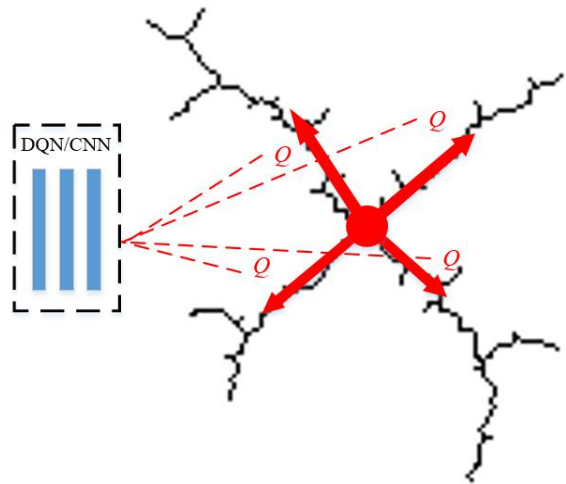

Figure 26. Fusion of DQN or CNN with RRT.

# VIII. Conclusion

This paper carefully analyzes principles of motion planning algorithms in section II-VI. These algorithms include traditional planning algorithms, supervised learning, optimal value RL and policy gradient RL. Direct comparisons of these algorithms are made in section VII according to their principles. Hence, a clear understanding about mechanisms of motion planning algorithms is provided. Analytical comparisons of these algorithms are made in section VII according to new criteria that include local or global planning, path length, optimal velocity, reaction speed, safe distance, and time-sequential path. Hence, general performances of these algorithms and their potential application domains are obtained. We specially compare the convergence speed and stability of optimal value RL and policy gradient RL in section VII because they are recent research focus on robotic motion planning. Hence, a detailed and clear understanding of these algorithms in network convergence are provided. Finally, we analyze a common motion planning task: long-distance motion planning with safety and efficiency (e.g. long-distance luggage delivery by robots) according to processing steps that include data collection, data preprocessing, motion planning and decision making. Hence, potential research directions are obtained, and we hope they are useful to pave ways for further improvements of motion planning algorithms or motion planning systems.